\documentclass{article}
\usepackage[preprint]{tmlr}
\usepackage[utf8]{inputenc}
\usepackage{microtype}
\usepackage{graphicx}
\usepackage{float}
\usepackage{booktabs}
\usepackage{array}
\usepackage{longtable}
\usepackage{enumitem}
\usepackage{varwidth}
\usepackage{afterpage}
\newsavebox{\codebox}
\usepackage{amsmath,amssymb}
\usepackage{url}
\usepackage{textcomp}
\graphicspath{{figures/}}

\title{Calibrated concept profiles: \\ making the reading of a model's neurons a measurement}
\author{\name Piotr Wilam \email piotrwilam@gmail.com \\
        \addr University College London}
\date{}

\begin{document}
\maketitle
\begin{abstract}
Interpretability has largely moved from reading a model's neurons directly to reading a second model trained on the first: a probe, a sparse autoencoder, a crosscoder. The move answered real faults --- mixed meanings, thresholds without a unit, readings that did not transfer from one study to the next --- but gave up what a direct reading alone provides: nothing trained in between, the neurons themselves named, and a reading that can be repeated and compared. Here the direct reading is made a measurement.

A neuron's activation varies with the prompt, so its reading is a distribution over a population of stimuli. We declare that population: every factor around a target construct is drawn at its rate in a reference corpus and logged.

From the declared distribution we generate a battery of prompts, and draw it twice. From the battery we compute the \emph{calibrated concept profile}, one number per neuron, stated in units of the neuron's own spread on ordinary input; the two draws give every number a noise unit. Before the profile is read, a fixed set of checks licenses the battery on the model, and the smallest change the profile can resolve across training checkpoints is measured.

Realised for Python on OpenCoder-1.5B and followed across checkpoints spanning two trillion training tokens, the profile gives what the alternatives do not report: a unit on every number, a chance level for every comparison, and which neurons carry a construct at every checkpoint.

The battery, the code and the data behind every result are released.
\end{abstract}

\section{Introduction}

\subsection{Why read neurons directly}

Reading a model's neurons directly is the oldest tool of interpretability, and the field has largely moved past it. The reasons are known. A single neuron can mix several meanings \citep{elhage2022}; a list of neurons has no stated threshold; and the readings did not travel from one study to the next. Learned decompositions --- linear probes, sparse autoencoders \citep{bricken2023,cunningham2023}, crosscoders \citep{lindsey2024} --- answered these by training a second model on the first and reading that instead. What is given up is not small: a direct reading has nothing trained in between, it names the model's own neurons, which can then be removed to test what they do, and it can be repeated --- on other stimuli, at another checkpoint, on another model of the same architecture --- and the results compared. This paper takes no side on whether neurons or learned features are the better unit. It claims that a direct reading can be made a measurement --- of a declared population of stimuli, in a stated unit, under a licence that says when it may be used, at a stated resolution --- and shows what such a measurement sees (\S4, \S5.1).

\subsection{What a reading is}

A neuron, as the word is used here, is one unit of a transformer's MLP block: at one position of the input it receives the residual stream, applies its weight vector and a nonlinearity, and outputs one number, its activation. Reading a concept from neurons means two things: a set of inputs shown to the model, the \emph{stimuli}, and a rule that turns the activations they produce into one statement per neuron, the \emph{arithmetic}. For a concept such as the \texttt{for} loop, the stimuli are prompts that contain \texttt{for} loops, and the arithmetic says, for each of the model's neurons, how much it responds to those loops. The question is the one asked since the first inspections of hidden units \citep{karpathy2016,bau2017}: which neurons respond to a given concept across many inputs, with whatever one prompt adds averaged out. The readings of named concepts in the literature --- knowledge neurons \citep{dai2022}, language-specific neurons \citep{tang2024}, universal neurons \citep{gurnee2024}, the neuron readings of code models \citep{sharma2023,kargaran2025,yin2025} --- all have this form.

\subsection{A reading is a distribution, over a population someone chose}

Two facts about the form decide what such a reading is. A neuron's activation is a continuous value, not a yes or no. And the stimuli vary: two prompts containing a \texttt{for} loop differ in their identifiers, their nesting, their length and in what stands near the loop, and the activation differs with them. So the reading of a neuron is a distribution: one value for each stimulus in some population, with a mean and a spread.

That population is the researcher's choice, and two ways of choosing it dominate. \emph{A template}, a handful of prompts written by hand: every factor is fixed, so the reading is conditional on choices nobody declared; circuits found on templated prompts behave differently off them, and sparsity does not remove the dependence \citep{wu2026,tan2024}. \emph{A sample from a corpus}, passages of real text that contain the construct: nothing is fixed, so the population cannot be described, it may overlap the model's training data, and there is nothing to subtract. The first is too rigid, the second too loose.

\begin{table}[t]
\centering
\scriptsize
\renewcommand{\arraystretch}{1.25}
\setcounter{table}{0}
\caption{The factors of a stimulus, and what each approach does with them.}
\label{tab:1}
\vspace{3pt}
\begin{tabular}{@{}>{\raggedright\arraybackslash}p{0.2500\dimexpr\textwidth-6\tabcolsep\relax}>{\raggedright\arraybackslash}p{0.2500\dimexpr\textwidth-6\tabcolsep\relax}>{\raggedright\arraybackslash}p{0.2500\dimexpr\textwidth-6\tabcolsep\relax}>{\raggedright\arraybackslash}p{0.2500\dimexpr\textwidth-6\tabcolsep\relax}@{}}
\toprule
factor of a stimulus & template & corpus sample & declared distribution \\
\midrule
the target construct in its role & fixed & present, uncontrolled & fixed (set A) \\\addlinespace[2.5pt]
the target's token out of role & rarely provided & uncontrolled & fixed (set B) \\\addlinespace[2.5pt]
identifiers and literals & fixed by the template & uncontrolled & drawn at corpus rates, logged \\\addlinespace[2.5pt]
scope and nesting & fixed & uncontrolled & drawn at corpus rates, logged \\\addlinespace[2.5pt]
companion constructs & fixed or absent & uncontrolled & drawn at corpus rates, matched between A and B, logged \\\addlinespace[2.5pt]
length and surrounding code & fixed & uncontrolled & drawn at corpus rates, logged \\\addlinespace[2.5pt]
ordinary behaviour of a neuron & not measured & not separated & measured on ordinary code (set C), per kind of position \\\addlinespace[2.5pt]
what a repeated draw gives & the same prompts & another sample & a second battery: the noise unit \\
\bottomrule
\end{tabular}
\end{table}

Table 1 sets the two ways against the declared distribution of \S2.1, factor by factor. The template column is the design of the minimal-pair benchmarks \citep{warstadt2020,gauthier2020}. The corpus-sample column is the design of studies that read neurons over passages of real text, such as the language-specific neurons of \citet{tang2024} and the universal neurons of \citet{gurnee2024}.

\subsection{What a measurement must state}

We take the position that, once the population is a choice, a reading counts as a measurement only if it states four things: the population, declared; the unit, the spread of the reading when the stimuli are drawn again from that population; evidence, from checks run before use, that its numbers come from the model's response and not from the stimuli or the arithmetic alone; and, for comparisons across training checkpoints, the smallest change it can resolve. Current neuron-level readings do not state them: thresholds without a unit, one checkpoint, and standardisation against pooled baselines although neurons are positional and token-class specific \citep{voita2024}. The variance of circuit discovery has been measured \citep{meloux2025,wu2026} and circuit components certified against resampling \citep{anani2026}, but not for neuron readings of named concepts.

\subsection{The solution}

The solution has two parts, one for each step of a reading.

\begin{enumerate}[topsep=2pt,itemsep=1pt,parsep=0pt,partopsep=0pt,leftmargin=*]
\item \emph{Controlled stimuli.} The prompts are generated, not collected or written by hand. A declared distribution lists every property around the target --- identifiers, nesting, length, the constructs nearby (Table 1) --- and draws each at the rate a reference corpus shows; the target is held fixed, everything else is as ordinary input of the domain has it, and every draw is logged, so that the population can be stated.
\item \emph{A reading that keeps the distribution.} A neuron's activations over the prompts are treated as a distribution, not as a value. The reading takes their mean, states it in units of the neuron's own spread on ordinary input, and reports how far it moves when the prompts are drawn again.
\end{enumerate}

We call the reading the \emph{calibrated concept profile} of a construct, the profile for short, and the method of making and licensing it goes by the same name.

\section{The calibrated concept profile}

The method is built in six steps (Figure 1). Each step answers a question that the previous step leaves open; \S3 realises the six for Python on OpenCoder-1.5B.

\begin{figure}[t]
\centering
\includegraphics[width=1\textwidth]{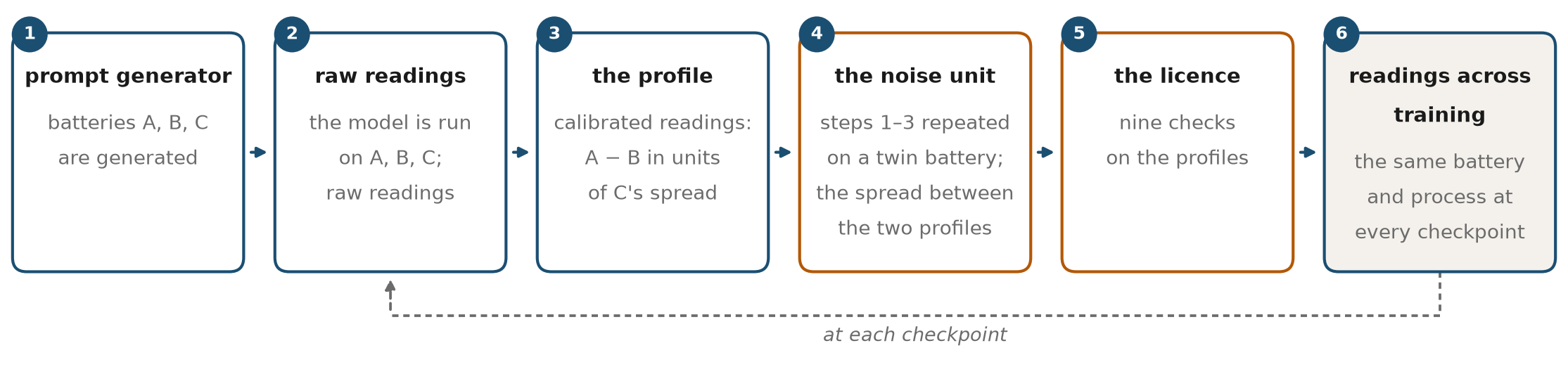}
\setcounter{figure}{0}
\caption{The calibrated concept profile in six steps (\S2.1--2.6), for one construct, at each checkpoint.}
\label{fig:1}
\end{figure}

\subsection{Step 1 -- prompt generator}

Which prompts should be shown to the model to read a construct, and how should they be chosen so that the reading can be repeated? A \emph{declared stimulus distribution} for a target is a written list of the factors around the target --- for code: identifiers and literals, scope and nesting, companion constructs, length and surrounding text --- and, for each, the rates at which its values are drawn, measured in a reference corpus. A prompt is one draw from the list; the draw is logged with the prompt. Nothing about a prompt is chosen for a reading. Three sets are drawn from the list, differing in what is held fixed:

\begin{enumerate}[topsep=2pt,itemsep=1pt,parsep=0pt,partopsep=0pt,leftmargin=*]
\item \emph{Set A} holds the target in its role and draws every other factor at its corpus rate.
\item \emph{Set B} holds the target's token outside its role --- inside a string, a comment, a docstring, or as part of a name --- and, where the token also has another grammatical role (the \texttt{for} of a comprehension, say), in that other role. Its companions are drawn so that their mix matches A's, in distribution rather than prompt by prompt, so that A and B differ only in the target's role; counterfactual datasets reach the same end one text at a time, by rewriting \citep{abraham2022}. B does for the reading what a control task does for a probe \citep{hewitt2019}: it removes what the token alone accounts for.
\item \emph{Set C} is the baseline: ordinary input of the domain, drawn at the corpus rates with no regard to any target. It is what a neuron's ordinary behaviour is measured on (\S2.3).
\end{enumerate}

Two properties of the declaration are used later: the logged draws can be re-weighted to another distribution by arithmetic, without running the model again; and the list can be drawn from twice (\S2.4).

\subsection{Step 2 -- raw readings}

A neuron's activation at the construct's position mixes several sources. How is the part due to the construct's role separated from the rest? The reading is made at one position per prompt, the \emph{read site}: the target's first token. A neuron's activation there has four sources. One is what we want, the target's role. Three are what we want to remove: the particular prompt; the token, which contributes the same whatever other role it is in; and the neuron's ordinary level and scale at that kind of position. Averaging over A removes the prompt: the mean activation per neuron over set A keeps what the prompts share, the role. Subtracting B removes the token: set B carries the same token and the same companions, read at its first token, so the mean over B holds what the token does on its own. The difference, mean over A minus mean over B, is the neuron's response to the role. Taken on the activations as the model produces them, we call it the \emph{raw form} of the profile. It is still in each neuron's own units: neurons differ in their ordinary level, one may sit near 0 and another near 3, and in their ordinary spread, one varying by tenths and another by tens, so a raw difference of 0.5 is large for one neuron and nothing for another. The next step restates it in each neuron's own spread.

\subsection{Step 3 -- the profile}

In what unit should a neuron's response be stated, so that neurons with different ordinary levels and spreads can be compared? In the neuron's own ordinary spread. For every neuron and every kind of position, $\mu$$_0$ and $\sigma$$_0$ are the mean and the spread of its activation over set C; an activation x becomes z = (x $-$ $\mu$$_0$)/$\sigma$$_0$. The pair ($\mu$$_0$, $\sigma$$_0$) per neuron and kind of position is the \emph{yardstick}, and the \emph{profile} in its standardised form is the difference of \S2.2 taken on z. An entry of 2 says: this neuron responds to the role by two of its ordinary spreads more than to the token alone.

The unit is taken per kind of position because a neuron's ordinary behaviour is not one number: its level and spread depend on where it is read. Take \texttt{len}, read at its name, in \texttt{n = len(x)} on the first line of a short prompt and in \texttt{for k in range(0, len(x)):} twenty lines in. The token is the same; the position is not, and a neuron's ordinary level and spread differ between the two, as they differ between a keyword and an identifier \citep{voita2024}. A single pair taken over all positions would make the unit wrong at both. A \emph{kind of position} is therefore a class of places in a prompt within which one pair is measured; which classes are distinguished is a choice of the realisation (\S3.3). One condition follows: A's and B's read sites must fall in the same kinds, so that $\mu$$_0$ cancels in the difference and $\sigma$$_0$ alone does the work.

Both forms of the profile are kept. The standardised form says which neurons carry the construct and how concentrated the response is; the raw form gives its size, because $\sigma$$_0$ itself changes over training and a standardised size would confound the two. Set C enters only through $\sigma$$_0$: it ties the unit to the reference corpus, which stands in for the model's training distribution (\S6).

\subsection{Step 4 -- the noise unit}

How much does the reading change when the prompts are drawn again from the same distribution, that is, how large is its noise? Every prompt is built from one of a few structurally distinct \emph{cores} of the target, dressed in drawn surroundings. The battery is drawn twice from the same list, on the same cores with different surroundings --- the \emph{twins}. Each twin gives a profile; their disagreement is what the surroundings alone can do. It is a noise floor in the sense of \citet{chen2026}, measured on the instrument before any effect is read against it. Because the twins share their cores this is surface noise only; variation between cores is carried by the bootstrap.

\begin{enumerate}[topsep=2pt,itemsep=1pt,parsep=0pt,partopsep=0pt,leftmargin=*]
\item \emph{The twin unit.} The distance between the two twins' profiles, d = $\surd$(2(1 $-$ cos)), where cos is the cosine between them. A difference between two profiles --- two constructs, two checkpoints --- is the same distance between them, stated in multiples of d.
\item \emph{The core bootstrap.} An interval on any number derived from a profile comes from resampling the cores with replacement and recomputing, with the same resampled cores for both twins and for every checkpoint compared.
\end{enumerate}

No distributional form is assumed: the yardstick is a change of unit, intervals come from the core bootstrap, and thresholds are stated in noise units, not as significance levels.

\subsection{Step 5 -- the licence}

Before a model is read, how do we know the instrument works on it? It may be used on a model when a fixed set of nine checks has run on the pair (battery, model): the two with a pass criterion have passed, and the other seven have been reported. A new battery, or a new model, needs a new licence. The checks measure the instrument, not the model's learning.

\begin{table}[t]
\centering
\scriptsize
\renewcommand{\arraystretch}{1.25}
\setcounter{table}{1}
\caption{The nine checks of the licence.}
\label{tab:2}
\vspace{3pt}
\begin{tabular}{@{}>{\raggedright\arraybackslash}p{0.1833\dimexpr\textwidth-4\tabcolsep\relax}>{\raggedright\arraybackslash}p{0.4167\dimexpr\textwidth-4\tabcolsep\relax}>{\raggedright\arraybackslash}p{0.4000\dimexpr\textwidth-4\tabcolsep\relax}@{}}
\toprule
check & what it asks & what its outcome restricts \\
\midrule
1. twin agreement & do two independently drawn batteries give the same profile? & any statement about which neurons carry the construct \\\addlinespace[2.5pt]
2. convergence & after how many prompts does the profile stop changing? & readings from fewer prompts than the convergence point \\\addlinespace[2.5pt]
3. length & does lengthening a prompt change the profile? & comparisons across prompts of different length, for the sensitive constructs \\\addlinespace[2.5pt]
4. re-weighting & does re-weighting the logged factors to other proportions change the profile? & comparisons across different distributions \\\addlinespace[2.5pt]
5. calibration & do the stated intervals cover the twin difference at their nominal rate? & the width of intervals: inflated by a stated factor \\\addlinespace[2.5pt]
6. companion & do A and B carry the same companion constructs, within a cap? & statements about the construct alone: its profile includes the companion difference \\\addlinespace[2.5pt]
7a. scrambled null & with A and B labels shuffled, does a profile remain? & a gate: the real profile must exceed the null's 95th percentile, or nothing is read \\\addlinespace[2.5pt]
7b. split-half null & do two halves of the same set give a "profile"? & a gate, as 7a \\\addlinespace[2.5pt]
8. yardstick floor & how far does the floor on $\sigma$$_0$ move the reading? & readings of neurons whose $\sigma$$_0$ sits at the floor \\
\bottomrule
\end{tabular}
\end{table}

The two nulls follow the corruption check of \citet{oikarinen2025} and the random-set null of \citet{kim2018}, applied to an instrument rather than to a metric. Checks 2, 3, 7a and 7b need the activation of each single prompt, since they add, pad or relabel prompts one at a time; the others work on the stored means.

The nine checks are of two kinds. The two nulls are a gate: if either fails, the profile cannot be told from what the arithmetic or the sampling produce alone, and nothing should be read. The other seven are not a gate. Each measures one way the reading depends on the particular battery, corpus or model --- on the draw (check 1), the number of prompts (2), prompt length (3), the mixture of logged factors (4), the stated intervals (5), the companions (6), the yardstick's floor (8). Their outcomes do not decide whether the instrument may be used; they state the restrictions within which it is used, and which experiments are safe to pursue within them is the researcher's decision. A profile that moves with the mixture is compared only over the same distribution; a length-sensitive construct is not compared across prompts of different lengths; a construct whose A and B differ in companions carries a companion part in its profile. The last column of Table 2 names the claims each outcome restricts. \S4.1 gives the restrictions measured on OpenCoder, and the claims of \S4 are made within them.

\subsection{Step 6 -- readings across training}

Which numbers are read from the profile, what are they compared with, to tell the model's response from what the battery and the architecture produce on their own, and how small a change between checkpoints can be detected? Three numbers, then one set and its ceiling:

\begin{enumerate}[topsep=2pt,itemsep=1pt,parsep=0pt,partopsep=0pt,leftmargin=*]
\item \emph{Norm} --- the size of the response, on the raw form.
\item \emph{Sparsity} --- the share of the squared norm in the largest 5\% of entries, on the standardised form: how concentrated the response is.
\item \emph{Twin agreement} --- the cosine between the two twins' profiles, from which the twin unit is computed; read as a property of the construct, how repeatable the response is.
\item \emph{The carrier set} --- the 5\% of neurons with the largest standardised entries. Two random 5\% sets overlap by 0.05.
\item \emph{The twin ceiling} --- the overlap of the two twins' carrier sets, the most two batteries can agree.
\end{enumerate}

The baseline is the same architecture with random weights, read with the same battery: whatever it gives is what the battery and the architecture give on their own. A construct is \emph{concentrated} if its sparsity exceeds the untrained value by more than three noise units and its twin agreement by at least 0.20; both margins were fixed before any reading. Two profiles of one architecture --- two checkpoints, two runs, a trained model and its untrained reference --- are compared in twin units: the distance between them, or the overlap of their carrier sets against the twin ceiling. This is what lets the method follow a construct through training.

A comparison across training checkpoints needs one more number: the \emph{resolution}, the step between a construct's profiles at neighbouring checkpoints, in twin units. Where it is below one, no date can fall between them.

\subsection{The choices the method rests on}

Which choices does every reading depend on, and how far is each dependence measured? Every reading is conditional on six choices. For each, Table 3 says what depends on it and how far this paper measures that dependence; \S2.5's checks are the measurements, and what is not measured marks the method's open edge.

\begin{table}[t]
\centering
\scriptsize
\renewcommand{\arraystretch}{1.25}
\setcounter{table}{2}
\caption{The six choices, what depends on them, and how far each is measured.}
\label{tab:3}
\vspace{3pt}
\begin{tabular}{@{}>{\raggedright\arraybackslash}p{0.3333\dimexpr\textwidth-4\tabcolsep\relax}>{\raggedright\arraybackslash}p{0.3333\dimexpr\textwidth-4\tabcolsep\relax}>{\raggedright\arraybackslash}p{0.3333\dimexpr\textwidth-4\tabcolsep\relax}@{}}
\toprule
choice & what depends on it & how far it is measured \\
\midrule
1. \emph{Two twins}: two draws of the battery, sharing their cores & every statement "beyond noise"; the twin ceiling; the resolution & twin agreement (check 1); intervals from the core bootstrap (\S4.8). Not measured: more draws, or draws on independent cores \\\addlinespace[2.5pt]
2. \emph{The generator}: cores and axes, menus of names, how B is built, companion matching & what the profile is a profile of & the mixture (check 4: median 2.34 twin units); companions (check 6: not matched for compound statements). Not measured: generated prompts against real code \\\addlinespace[2.5pt]
3. \emph{The yardstick}: token class $\times$ fifth of the prompt; set C; the floor on $\sigma$$_0$ & which neurons are named; size in standardised units & coarser yardsticks name different neurons (\S4.3); the floor (check 8); length (check 3). Not measured: other splits, computed exactly \\\addlinespace[2.5pt]
4. \emph{The reference corpus}: public Python, not the training data & what "ordinary" and "at corpus rate" mean & not measured: how far public Python differs from the model's 607-language mixture \\\addlinespace[2.5pt]
5. \emph{Naming rules}: the inventory, the read site, the gate, the top-5\% carrier set & which constructs are read; which neurons are the carriers & the gate's failures (19 of 188, \S3.1); the cut does not change carrier turnover \citep{wilam2026}. Not measured: another read site \\\addlinespace[2.5pt]
6. \emph{The read point}: MLP neurons, one position, short prompts (median 55 tokens) & everything read & not measured: longer contexts, attention, the residual stream \\
\bottomrule
\end{tabular}
\end{table}

\section{One realisation: Python on OpenCoder-1.5B}

Three things were built, each from the one before: a battery of prompts, a record of the model's activations on it, and the calibrated concept profiles computed from the record. Each choice is one way of meeting \S2; the generator's mechanics and the details are in Appendix A.

\subsection{The battery}

\emph{The inventory: the whole language.} The constructs are not a chosen sample: every node of Python 3.12's grammar that appears as text of its own, and every public builtin name --- 63 + 143 = 206 constructs, of which 188 were built. What is excluded is excluded by a rule applied to the whole grammar, not by a judgement about any particular construct (Appendix A.1).

\emph{The battery.} OpenCoder's training data was not released, so a public corpus of 253,398 parsed Python files is the reference corpus, and its rates are the declared rates. Each construct has eight \emph{cores} drawn from a family of between 42 and 1,280 (median 210), and each of A and B holds 720 prompts in each twin, ninety from every core, the same for all 188 constructs --- 1,440 prompts a set across the two twins; companions are drawn at corpus rates and matched between A and B; every prompt is formatted, parsed and logged. For \texttt{for}, A's prompts are loops --- \texttt{for x in range(n):} with a body, among drawn identifiers, at a drawn scope, with the companion statements real code puts near a loop --- and B's carry the token \texttt{for} inside a string, a comment or a name --- with the same companions; for the eighteen constructs whose token has another role, B also holds it in that role (Appendix A.2). The generator never runs the model and never reads the training data.

\emph{The read site and its gate.} The read site is the first token of the construct's node: on \texttt{for x in range(n):} the token \texttt{for}; for a call to \texttt{len} the name \texttt{len}. Every recorded quantity carries the prompt's scope, because a statement-initial keyword is a different token indented (\texttt{\textvisiblespace{}for}) and at module level (\texttt{for}), two embeddings as unrelated as a random pair (cosine 0.056). A \emph{read-site gate} admits a construct only if the most common token piece at its read site covers at least 80\% of prompts, counted on the indented scopes, in both twins and on both A and B: 169 of 188 pass; these form P$_1$, the population read in \S4. The nineteen failures are listed in Appendix A.4. The gate is the one place where the tokenizer, not the grammar, decides what can be read.

\emph{Artefact:} the battery in full --- the prompts of A and B for all 188 constructs in both twins, the set C, set A regenerated for every construct from a second declared distribution, sixty prompts each --- check 4's alternative menu, not a further draw from the same menu, which is what a twin is --- and the log of every draw --- with its hash (\texttt{570b42db\ldots{}}), in the dataset \texttt{piotrwilam/AtlasOpenCoder}, tag \texttt{v1}.

\subsection{The record}

\emph{The model and its neurons.} OpenCoder-1.5B \citep{huang2024} has 24 layers with an MLP hidden width of 6,144 (SwiGLU). Each MLP block's hidden layer has 6,144 units; these are the neurons. A unit's activation is the product of its gate branch, after the nonlinearity, and its up branch, taken before the down-projection returns it to the residual stream (Figure A.2); it is the quantity \citet{geva2021} read as a key-value memory. All 24 $\times$ 6,144 = 147,456 neurons together are the space in which comparisons across checkpoints are made, which we call the \emph{product space}.

\emph{The references.} Sixty-one references were shown the battery: 35 pretraining checkpoints (the twenty from 4.19 to 83.9 billion tokens, then fifteen spaced $\times$1.25 to 2.03 trillion), 24 annealing checkpoints and two untrained references. The per-core means are stored, with the cell axis kept, so that the size of the carrier set can be changed later, and the yardstick by merging cells, without re-recording (Appendix A.6).

\emph{What is stored per prompt.} The per-core means are enough for every reading in \S4 and for checks 1, 4, 5 and 8. Single-prompt activations, which checks 2, 3, 7a and 7b and the probe of \S4.7 need, are kept at four references --- the untrained model, 1,000 and 20,000 steps, and the final checkpoint: the MLP neurons for 31 constructs drawn to represent the read population, and the residual stream for 164 (Appendix A.6). Check 6 applies to the eight constructs that can have a companion: \texttt{ClassDef}, \texttt{For}, \texttt{FunctionDef}, \texttt{If}, \texttt{Match}, \texttt{Try}, \texttt{While}, \texttt{With}.

\emph{Artefact:} the record is not released in full; at 27.6 GB of accumulators a reference it comes to 2,022 GB. The dataset holds a worked example --- twin 1's accumulators for \texttt{for}'s A and B sets and for set C at the final checkpoint (272 MB) --- and the recorder in the repository records any reference again from the model's public checkpoints and the battery.

\subsection{The profiles}

\emph{The cells.} A kind of position (\S2.3) is here a \emph{cell}, defined by two properties of a position: the class of the token at it --- keyword, identifier, symbol, newline or indent --- and the band it stands in, which fifth of the prompt's length. Four classes times five bands gives 20 cells, and 20 $\times$ 147,456 = 2.95 million ($\mu$$_0$, $\sigma$$_0$) pairs per reference. The token class is the property a neuron's ordinary behaviour depends on (\citealp{voita2024}; \S2.3). The band is relative to the prompt's length rather than absolute for two reasons: every cell of C is then populated across prompts of 2 to 313 tokens, and A's and B's read sites fall in the same cells, as \S2.3 requires. The cost is that the same prompt, lengthened, can move its read site to a different cell (\S4.1).

\emph{From the record to a profile.} (1) On set C's part of the record, for each neuron and each of the 20 cells, the mean $\mu$$_0$ and spread $\sigma$$_0$ of the activations at tokens in that cell: the yardstick, which knows nothing about any construct. (2) For each A prompt of a construct, the read site's cell is found from its token class and band --- for a \texttt{for} loop at an indented scope, a keyword in the band of the loop's line --- and each neuron's activation there is standardised with that cell's pair. (3) The same for each B prompt. (4) For each neuron, the mean over A minus the mean over B: the profile, one number per neuron; the same on the raw activations gives the raw form. The standardisation is applied to each prompt's activation before averaging, with the pair of that prompt's own cell; the two forms are therefore two different averages, not one average rescaled. (5) Steps 2--4 run on both twins; from the two profiles come the twin unit, the carrier set (7,373 neurons) and its twin ceiling, and the core bootstrap's intervals (\S2.4, \S2.6).

\emph{Artefact:} the carrier sets of every construct at all 61 references, with the twin ceilings and the licence tables of Appendix B; and, as a worked example, the profile of \texttt{for} in twin 1 at the final checkpoint, every array. The other profiles follow from the record.

\section{The instance: what the battery on OpenCoder shows}

\S1.4 said what a reading must state to count as a measurement. The population is \S3.1's. \S4.1--4.4 state the other three for this instance: the evidence that the numbers are the model's (the licence, \S4.1; the ablation, \S4.4), the resolution (\S4.2) and the unit's load (\S4.3); \S4.5--4.8 show what the instrument then sees.

\subsection{The licence as run}

Does the instrument work on OpenCoder-1.5B --- do its nulls pass, and how stable, convergent and calibrated is its reading? On the terms of Table 4: both nulls pass with a wide margin for every construct they ran on, so the gate is passed; the seven other checks show a stable, convergent reading, calibrated at the median, with three restrictions, from checks 3, 4 and 6, stated below and in the licence sentence.

\begingroup
\scriptsize
\renewcommand{\arraystretch}{1.25}
\setlength{\LTcapwidth}{\textwidth}
\setcounter{table}{3}
\begin{longtable}{@{}>{\raggedright\arraybackslash}p{14pt}>{\raggedright\arraybackslash}p{0.1667\dimexpr\textwidth-6\tabcolsep-14pt\relax}>{\raggedright\arraybackslash}p{0.0667\dimexpr\textwidth-6\tabcolsep-14pt\relax}>{\raggedright\arraybackslash}p{0.7667\dimexpr\textwidth-6\tabcolsep-14pt\relax}@{}}
\caption{The licence on OpenCoder-1.5B, final checkpoint.}\label{tab:4}\\
\toprule
\# & check & ran on & outcome \\
\midrule
\endfirsthead
\toprule
\# & check & ran on & outcome \\
\midrule
\endhead
\midrule
\multicolumn{4}{r}{\itshape continued} \\
\endfoot
\bottomrule
\endlastfoot
1 & twin agreement & 188 & median cosine 0.986 between the twins' profiles \\\addlinespace[2.5pt]
2 & convergence & 31 & cosine 0.99 with the full set reached by 56 of 720 prompts (median) \\\addlinespace[2.5pt]
3 & length & 31 & a median 2.4\% of a construct's neurons length-sensitive; more than 5\% for eight statements (\texttt{With} 55\%, \texttt{For} 49\%) \\\addlinespace[2.5pt]
4 & re-weighting & 188 & designed arm only; the re-weighted profile sits a median 2.34 twin units from the base, within one twin unit for 4 of 188 --- the profile moves with the mixture \\\addlinespace[2.5pt]
5 & calibration & 188 & calibrated at the median; intervals optimistic for 33 constructs, by a stated factor \\\addlinespace[2.5pt]
6 & companion & 8 & the generator's matching is not met: the largest difference between A's and B's share of a companion is 0.19--0.68 (median 0.37) against the design's cap of 0.15; removing the difference, where it can be removed, moves the profile by 2.95 twin units (\texttt{For}) and 5.50 (\texttt{With}) \\\addlinespace[2.5pt]
7a & scrambled null & 31 & real profile a median 10.7$\times$ the null's 95th percentile (minimum 7.2$\times$); 0 of 31 at or below --- pass \\\addlinespace[2.5pt]
7b & split-half null & 31 & median 12.8$\times$ (minimum 6.0$\times$); 0 of 31 at or below --- pass \\\addlinespace[2.5pt]
8 & yardstick floor & 188 & moves a profile by a median 0.0039 and at most 0.035 twin units \\
\end{longtable}
\endgroup

Check 4 is the reason the distribution is declared (\S2.5): re-weighting the logged factors moves the profile by more than twice what redrawing the battery does. Length sensitivity is partly the yardstick's doing (Appendix B.4); it does not affect comparisons across checkpoints, because every checkpoint sees the same prompts. Check 6 restricts what can be said about the eight compound statements: their B sets do not carry the same companion constructs as A (Table B.3), so their profiles include a companion part. Where B can be re-weighted to A's companion shares --- \texttt{For} and \texttt{With} --- the profile moves by 2.95 and 5.50 twin units, of the order of check 4's re-weighting. The companion part does not change the turnover: between every pair of the four per-prompt references, the carrier overlap of \texttt{For} and \texttt{With} moves by at most 0.008 when B is re-weighted, so comparisons across checkpoints are not affected. It changes which neurons are named for \texttt{With} but not for \texttt{For}, and the re-weighting is not adopted as the reported profile (Appendix B.3). Per-construct values are in Appendix B.

\emph{The licence sentence:} licensed on the gate --- the two nulls passed on the 31 constructs they can run on --- with these restrictions: the profile moves with the mixture by a median 2.34 twin units (check 4, designed arm only); eight statements are length-sensitive (check 3); for the eight compound statements A and B differ in companions, by 2.95 and 5.50 twin units where measurable (check 6); 33 constructs' intervals are optimistic by a stated factor (check 5); the second-model replication outstanding and named.

\subsection{The resolution, and a construct followed through training}

How small a change in a construct's profile between two checkpoints can the record detect, and so how finely can it date a change? To the spacing of the checkpoints, 1,000 training steps: even at that spacing the step exceeds one twin unit, so the instrument's own limit lies below it. On P$_1$ the median step between neighbouring references is 8.56 twin units between the first two checkpoints (1,000 steps apart) and 3.56 at the end of the dense block; 4.06 across the late tail and 2.91 on the annealing branch. The minimum is above one everywhere except \texttt{None}, at 0.862 and 0.918 in two pairs. At the same 1,000-step spacing early training moves the profile further than late: 8.56 twin units against 3.56, a ratio of 2.4.

The same profile, read at 61 references, follows one construct from the first checkpoint to the end. For \texttt{for}, the raw norm ends at 52\% of its first-checkpoint value while the standardised norm does not fall, which is why size is read on the raw form (\S2.3, \S4.8; Appendix C.2). \citet{wilam2026} reads the whole record this way, for every construct, at this resolution.

\subsection{The unit is load-bearing}

Does the choice of yardstick change which neurons are named? Yes: the yardstick is part of the measurement. Change what counts as ordinary at a position and the named neurons change by more than redrawing the battery changes them. Under coarser yardsticks the carrier sets overlap the 20-cell sets by a median 0.73--0.87, below the twin ceiling for 78 to 167 of the 169 constructs, and the token class carries more of the set than the position band. A comparison with an alternative split, registered in the design, is deferred: the stored record, kept in the 20 cells, supports a split that cuts across them exactly for only the 31 constructs with per-prompt activations, so it needs a new recording keyed by the alternative's cells.

The unit also separates the model from the battery. Twin agreement, concept norm and carrier overlap with the previous reference are 0.981--0.989, 48.7--68.8 and 0.822--0.851 on every trained reference, and 0.710 / 6.34 / 0.273 and 0.717 / 6.13 / 0.286 on the two untrained references, recorded days apart under different seeds; the twin ceiling rises from 0.28 untrained to 0.84 at the first checkpoint (plug-in; core-bootstrap means 0.21 $\rightarrow$ 0.74).

\subsection{The named neurons carry function}

Do the neurons the profile names as carriers affect the model's predictions at the construct's positions? Removing them raises the loss where the construct's grammar fixes the next token. For the 16 constructs whose prompts carry such positions, zeroing the carrier set raises the loss there by more than zeroing a random set of the same size does: by a median 0.06 nats at 4.19 B tokens [0.02, 0.14], 0.25 at 83.9 B [0.03, 0.66] and 0.40 at 2,034 B [0.10, 0.70] (construct-bootstrap 95\% intervals over 200 random sets; above the random mean for 12, 14 and 14 of 16 constructs, sign test p = 0.038, 0.002, 0.002). The largest: \texttt{Try}, whose carriers cost 5.95 nats more than a random set of the same size, on a 0.088-nat baseline; \texttt{Lambda}, 4.99 nats more, on 1.40. The comparison is with the random sets' mean and by a sign test over constructs, not in units of the random sets' spread, which is long-tailed (Appendix C.1; Table C.1 gives the same comparison under one pooled cell). At 4.19 B the median effect is a fifth of its value at 83.9 B: the neurons acquire most of their function during the first 20$\times$ of tokens.

\subsection{The model's \texttt{for} is neither a token nor a role}

Do the neurons that carry \texttt{for} follow the token \texttt{for}, whatever its role, or the role of a loop, whatever its token? Neither: two roles of one token share a quarter of their neurons, and one role across two tokens shares half. \emph{One token, two roles:} \texttt{for} as loop and \texttt{for} as comprehension, separate A sets with their own cores, have carrier sets overlapping by 0.250--0.266 at the last checkpoint (0.198 for \texttt{AsyncFor}); chance is 0.050 and the ceiling 0.872, so five times chance and under a third of the ceiling. \texttt{if} as statement and as conditional expression likewise, 0.266. \emph{One role, two tokens:} \texttt{\textvisiblespace{}for} and \texttt{for}, the loop indented and at module level, overlap by 0.494 against the two scope-restricted sets' own twin ceilings of 0.837 (module) and 0.850 (indented), lower than the whole construct's 0.872 because each scope's set is cut from fewer prompts: 0.59 of the lower ceiling, ten times chance. Across the twelve constructs whose token differs by scope the share runs from 0.44 (\texttt{Match}) to 0.71 (\texttt{Try}), and at the first checkpoint every share is lower: the two tokens' carrier sets converge over training without becoming one. Figure 2 shows where the carriers sit: ordered by how many of the 169 carrier sets a neuron belongs to, they concentrate on the neurons that carry many constructs, up to five times the 5\% rate, and thin out towards neurons that carry few without stopping; and they sit in every one of the 24 layers. The full map of all 147,456 neurons, with the two scopes' sets drawn on it, is Figure C.1.

\begin{figure}[t]
\centering
\includegraphics[width=1\textwidth]{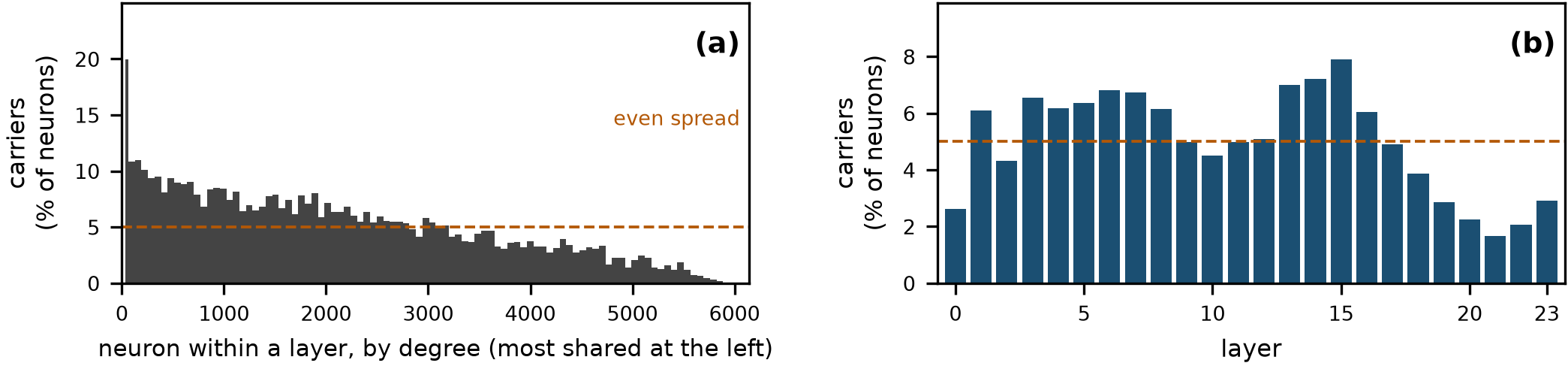}
\setcounter{figure}{1}
\caption{Where the \texttt{for} loop's carriers sit, at the end of pretraining (2,034 B tokens). (a) Carriers per unit of neurons, the neurons ordered by degree, the number of the 169 carrier sets each belongs to (high degree at the left); the dashed line is the 5\% rate carriers would have if spread evenly. (b) Carriers per layer, against the 5\% rate. The carrier set is concentrated on high-degree neurons and present in every layer.}
\label{fig:2}
\end{figure}

\subsection{Shared neurons follow the inventory, not the grammar}

When two constructs share carrier neurons, what do they have in common --- a category of the inventory, or a place in the grammar? Their category in the inventory, not their place in the grammar. The carrier-set overlap of P$_1$'s 14,196 pairs at the last checkpoint, clustered at 2 to 32 clusters and each inventory category scored for purity against 1,000 label shuffles, recovers grammar element against builtin name, the builtins' exception, warning, function and type classes, and one grammatical family, the comprehensions (purities and p-values in Table C.2). It does not recover statement against expression, compound statements, binding statements, or composites and their obliged parts. The highest-sharing pairs across categories pair a statement with the builtin that does the same job (\texttt{Delete}--\texttt{delattr}, \texttt{Global}--\texttt{globals}). The result does not come from the yardstick: with every carrier set rebuilt under a single pooled cell, every category that clusters significantly is still an inventory category and none a grammatical one (Appendix C.4).

\subsection{A probe's accuracy saturates; the profile does not}

Can a linear probe, the field's default instrument, track a construct across checkpoints as the profile does? On a trained checkpoint it cannot, because it separates everything. A linear probe on one layer at the read site --- residual stream for 164 constructs, hidden units for 31, folds holding out whole cores, shuffled-label controls at 0.50 --- separates A from B on the untrained reference only partly: median accuracy 0.653 on the residual stream, 0.802 on the hidden units, and no construct reaches 0.99. On every trained checkpoint it separates them almost completely: median 0.998 at the first checkpoint, 0.999 at 84 B and at the end; 145--148 of 164 constructs at 0.99 or above; minimum 0.902. Accuracies span 0.902 to 1.000 with most at the ceiling, which cannot order 169 constructs, 24 layers or 61 checkpoints; the area under the curve is no better, and a harder contrast --- two roles of one token --- saturates too (Appendix C.5). The probe's margin, its one continuous reading, does move --- a median 13.8 at 4.19 B, 16.5 at 83.9 B and 14.8 at the end (Appendix C.5) --- but it is one number per construct and names no neurons; on the same two-role pairs the profile discriminates (\S4.5), and its numbers move over training.

\subsection{Three artefacts of the instrument, and the rules they produced}

Which readings of this instance were wrong, and what rule replaced each? Three were, and each has been measured and is governed by a stated rule. Each is a trap for any reading of neurons across checkpoints, not only this one.
\begin{enumerate}[topsep=2pt,itemsep=1pt,parsep=0pt,partopsep=0pt,leftmargin=*]
\item \emph{The twin unit is too small for intervals.} The two twins share their cores, so their disagreement measures the variation of surface details (names, literals) and not the variation from choosing other cores; agreement corrected by the twin unit comes out above 1 for about 150 of 188 constructs. Rule: intervals come from the core bootstrap.
\item \emph{The standardised size grows while the real size shrinks.} The yardstick's spread $\sigma$$_0$ shrinks over training, so the standardised norm rises for 144 of 169 constructs while the raw norm falls for 148. Rule: size is read in raw units.
\item \emph{The layer of largest response is unstable.} Excluding the last layer moves it for 41\% of constructs. Rule: comparisons are made in the product space, over all 147,456 neurons together, not layer by layer.
\end{enumerate}

\section{Where the profile stands}

\subsection{Against the alternative instruments}

\textbf{The situation.} A researcher wants to know which neurons of a model carry a named construct, how reliably, and how that changes across hundreds of training checkpoints. The comparison below is limited to that situation. Outside it, several of these methods do things the calibrated concept profile cannot.

\emph{Probes.} A probe is a classifier trained to read a property from the model's activations. It gives a direction in activation space and an accuracy. It does not say which neurons carry the property, it gives no chance level for comparing two constructs' carriers, and it is not causal; following a construct across checkpoints needs one probe per checkpoint. The whitened profile --- the profile multiplied by the inverse covariance of the activations --- is the mass-mean probe direction \citep{marks2023}, so a probe of this kind is one view of the profile.

\emph{Sparse autoencoders and crosscoders.} A sparse autoencoder is a second model trained to rewrite the activations as a sparse sum of learned features. It is trained on one checkpoint's activations. Features of two checkpoints are matched only by a crosscoder trained across them \citep{bayazit2026}. The stimuli are the training corpus, so they are not declared, and a feature's activation has no noise floor; a feature is tested causally only by steering, as a separate step. Its strength is the one the profile lacks: it finds concepts nobody named. The two are complements. The profile tests a named concept with a unit; the dictionary proposes concepts to test.

\emph{Attribution and patching.} These methods change one component's activation on one prompt and measure the effect on the output. They are causal, and they trace paths between components. They work per prompt, so they give no population statistic, and no unit for comparing checkpoints. The ablation of \S4.4 applies their standard to the profile: the named neurons are removed and the loss at the construct's positions is read.

\emph{Neuron lists.} Knowledge neurons are the neurons with the largest integrated-gradient attribution for a fact \citep{dai2022}; language neurons are the neurons whose activation probability is most specific to one language \citep{tang2024}. They are the profile's nearest relatives. They differ in three ways: the threshold has no unit, the reading is made at one checkpoint, and a neuron's ordinary behaviour at the position read is not removed; they are also rarely tested causally. The twin unit (\S2.4), the yardstick (\S2.3) and the licence (\S2.5) are the additions. The turnover result of \citet{wilam2026} shows why they matter: a list read at one checkpoint describes that checkpoint.

\emph{Circuit-stability methods.} \citet{anani2026} certify each component's inclusion in a circuit as stable under bounded edits of the concept dataset, by randomised subsampling. \citet{wu2026} measure how much a circuit changes when the prompts are resampled and when they are rephrased with other templates, and find that sparsity does not remove the template dependence. Their resampling and rephrasing variances are the counterparts of our core bootstrap and twins. Both certify circuits, made of components and edges, in trained models at one checkpoint. The profile certifies neurons, at every checkpoint.

\textbf{What the profile adds, and what it does not claim.} The calibrated concept profile gives, for a named construct, a reading of the model's neurons that is a measurement in the sense of \S1.4: it states the four things a measurement must state.
\begin{enumerate}[topsep=2pt,itemsep=1pt,parsep=0pt,partopsep=0pt,leftmargin=*]
\item \emph{The population}, declared (\S2.1).
\item \emph{The unit}: the reading's spread when the battery is drawn again, the noise unit on every number (\S2.4); each neuron's response is first stated in that neuron's own spread at the position read (\S2.3).
\item \emph{Evidence that its numbers come from the model}, from a licence run before any reading (\S2.5, \S4.1).
\item \emph{For comparisons across checkpoints, the smallest change it can resolve}: the same units at every checkpoint and a measured resolution of 1,000 training steps, so one reading follows a construct through two trillion tokens (\S4.2).
\end{enumerate}

None of the other methods gives all four for a reading of neurons, and the profile trains nothing in between.

We read neurons, not features, because neurons are what training moves and the only units that can be followed across checkpoints without a second trained model. The carrier set is a measurement, not a claim that neurons are concepts.

\section{Limitations}

\emph{Of the method.} The six choices the calibrated concept profile rests on, and how far each dependence is measured, are in \S2.7 (Table 3); two are not measured at all --- the reference corpus and the read point. Beyond them: it reads named constructs only, so it cannot discover concepts; and everything it says is relative to its distribution, so a neuron in no construct's carrier set may carry something the battery does not contain.

\emph{What changes in another setting.} The method is not specific to code. The A/B/C design, the twins, the yardstick's principle (each neuron in its own spread at the same kind of position), the licence and the resolution do not depend on Python or on this model. What a new setting must supply: an inventory of named constructs fixed outside the model; a parser that decides a construct's role, or for natural language a tagger; a definition of the token out of its role, for set B (per word, in natural language); a reference corpus for the declared rates and for set C; a choice of cells for the yardstick, licensed again; and a read site, which for a model without MLP neurons (a state-space model, say) must be redefined. Everything in \S3 is one realisation of these.

\emph{Of this instance.} What the choices of Table 3 cost on OpenCoder --- the gate, length sensitivity, the mixture, the companions --- is stated in \S3.1 and in \S4.1's licence sentence. Beyond them: one model, one language --- the battery is Python on a model that read 607 programming languages, so a construct's exposure in training is larger than its Python rate, and nothing here is shown on a second model. A battery that matches companions, and the natural arm of check 4, are what a new recording would add.

\bibliographystyle{tmlr}
\bibliography{refs}
\appendix
\section{The realisation in full}

\subsection{The inventory and the reference corpus}

\begin{enumerate}[topsep=2pt,itemsep=1pt,parsep=0pt,partopsep=0pt,leftmargin=*]
\item \emph{The inventory.} Every entry of Python 3.12's grammar that appears as text, and every builtin name: 63 + 143 = 206 constructs. 188 were built; 18 were not (item 3).
\item \emph{What the rule excludes.} Of Python 3.12's grammar nodes, 48 never appear as text of their own, and are left out: module containers, context markers, pure wrappers, product types, and the operator markers that fold into \texttt{BinOp}, \texttt{BoolOp}, \texttt{Compare} and \texttt{UnaryOp}. That leaves 63. The builtins are every public name in \texttt{builtins} except the six the interactive shell injects (\texttt{help}, \texttt{exit}, \texttt{quit}, \texttt{copyright}, \texttt{credits}, \texttt{license}).
\item \emph{The eighteen not built.} Seven have a corpus rate but admit no family of cores, because several grammar nodes share one surface token (\texttt{None} three times, \texttt{yield} twice, and \texttt{await}, \texttt{\{}, \texttt{f"}). The other eleven are the syntax Python added in versions 3.10--3.12 --- the six \texttt{match} patterns, \texttt{except*}, and the three type parameters and the type alias of PEP 695 --- which the reference corpus does not contain often enough to calibrate.
\item \emph{The reference corpus.} 253,398 parsed Python files. The written list of factors and rates drawn from it is $\nu$, and $\nu$'s rule is that every property of a prompt is set to the corpus rate, never chosen for a reading.
\end{enumerate}

\subsection{Sets, cores, twins, scope}
\begin{enumerate}[topsep=2pt,itemsep=1pt,parsep=0pt,partopsep=0pt,leftmargin=*]
\item \emph{Sets A and B.} 720 prompts per set in each twin, ninety from each of eight cores, for every one of the 188 constructs; a set across both twins is 1,440 prompts.
\item \emph{Cores.} Drawn from a family whose size is the product of the construct's axes, from 42 (\texttt{Starred}) to 1,280 (\texttt{For}), median 210 (\texttt{CORE\_COUNTS.md}).
\item \emph{B's sub-sets.} One per kind of out-of-role use --- docstring, comment, string, identifier --- and, for the 18 constructs whose token is shared with another construct (\texttt{for}: \texttt{For}, \texttt{AsyncFor} and the four comprehensions; \texttt{def}, \texttt{with}, \texttt{if}, \texttt{:}, \texttt{\{}, \texttt{[}), one with the token in the other construct's role, built from that construct's own cores: 4,850 prompts, 1.8\% of B (\texttt{B\_SUBTYPES.md}). No rendered prompt appears in both a construct's A set and another construct's B set.
\item \emph{Set C.} One pool for all constructs, generated from the same menus and the same formatter as A and B with no construct anchored in it, and longer: 97.7 tokens on average against A's 56.9.
\item \emph{Scope.} Every recorded quantity carries the prompt's scope --- module, function or method --- because one construct is read through two different tokens depending on where it stands (\S3.1, \S4.5).
\item \emph{In all.} 569,270 prompts and 33,271,350 tokens, 58.45 tokens per prompt (median 55, quartiles 28--81).
\end{enumerate}

\subsection{How one prompt is made}

Four stages, each fixing one thing. (1) \emph{The core fixes the construct's own shape.} A core is one combination of the construct's axes; for \texttt{for} the axes are the operand, the suite, the access and the body, and the family is every combination, 1,280. Eight are drawn. Core c3 of \texttt{for} is \{operand: attribute, suite: two statements, access: bare, body: early exit\}; core c2 is \{operand: \texttt{range}, suite: companion, access: subscript, body: count\}. (2) \emph{The cell fixes the surroundings.} A cell is (domain, scope, padding, draw): five vocabulary domains (finance, biology, gaming, physics, e-commerce) $\times$ three scopes (module, function, method) $\times$ three padding levels $\times$ two draws = 90; every cell appears under every core, so 8 $\times$ 90 = 720 prompts a set. (3) \emph{The menu fills the names.} Identifiers and literals are drawn at corpus rates and injected into the slots the core declares; the core never chooses a word. The same core and cell drawn twice give the same shape with every name new (Figure A.1):

\begin{lrbox}{\codebox}
\begin{varwidth}{\maxdimen}
\begin{verbatim}
# For/c3, module scope, e-commerce, draw 1
requirement_val = next_indent_head.run_command
error_level_ok  = repo_name_total > 0
for threading_list in ParseNolintSuppressions_store.ParseNolintSuppressions:

# For/c3, draw 2
coupon_redemption_val = STUDENT_VIEW_head.NumLines
if carriage_return_found_ok:
    PrintUsage_all(commentpos)
for newpath in closing_brace_pos_store.closing_brace_pos:

# For/c2, module scope, biology
click_val = overlap_head.statements
for iterator in range(3):
    if ligand_ok:
        transcript_total += iterator * 1
\end{verbatim}
\end{varwidth}
\end{lrbox}
\par\noindent\hspace*{0.06\columnwidth}\resizebox{0.8\columnwidth}{!}{\usebox{\codebox}}\par

\vspace{4pt}
\noindent Figure A.1: Three prompts from the \texttt{for} battery: core c3 drawn twice (module scope, e-commerce), and core c2 (module scope, biology). The shape is the core's; every name is the menu's.
\vspace{12pt}

(4) \emph{Set B.} B's prompts are built from their own shapes, which place the token in a docstring, a comment, a string or an identifier --- and, for the 18 token-sharing constructs, in the sibling construct's role (for \texttt{for}: \texttt{[(d, 2) for k, d in zip(\ldots{})]}).

(5) \emph{The companion cap.} B's companion rates are fitted so that B's companion mix matches A's; the cap is the check that it does.

(6) \emph{Format, parse, read site, log.} Every prompt is formatted with \texttt{black} and parsed, and its construct's node is located. The read site is the node's first token. Every draw is logged with the prompt. The generator's own rule, in its source: "data in the loop is fine; the measured model in the loop is never".

(7) \emph{One asymmetry between A and B.} A's grid is balanced by construction: 240 prompts per scope and 240 per padding level. B's is not: for \texttt{for}, 219/234/267 by scope and 175/262/283 by padding, because B's shapes do not all admit every cell. The yardstick standardises each read site in its own cell, so the imbalance does not enter the profile; a cell-by-cell comparison of A and B should expect it.

\subsection{The read-site gate}

A construct is read only if its A and B prompts land on the same token piece in at least 80\% of prompts, in both twins, on the indented scopes. Named builtins: 143 pass, none fail. Structural constructs: 1 passes, 14 fail --- the operator families, the symbol-anchored expressions and \texttt{Assign}, \texttt{AnnAssign} --- because the tokenizer merges a symbol with its neighbours. Keyword tokens: 25 pass, 5 fail because the piece at the read site is not one piece --- \texttt{FunctionDef}, \texttt{ClassDef}, \texttt{AsyncFunctionDef} split \texttt{\textvisiblespace{}def}/\texttt{def} (\texttt{\textvisiblespace{}class}/\texttt{class}) across B's subtypes, \texttt{BoolOp} reads \texttt{\textvisiblespace{}and} or \texttt{\textvisiblespace{}or}, \texttt{UnaryOp} reads \texttt{\textvisiblespace{}not}, \texttt{\textvisiblespace{}\textasciitilde{}} or \texttt{\textvisiblespace{}-}. The gate is the realisation's dependence on its tokenizer, and the first thing a port to another model must re-check.

\subsection{The neuron that is read (Figure A.2)}

One MLP block of OpenCoder-1.5B. The residual stream (width 2,240) enters two trained projections, the gate (2,240 $\times$ 6,144) followed by the SiLU nonlinearity, and the up (2,240 $\times$ 6,144); their outputs are multiplied unit by unit, giving 6,144 numbers, the neurons' activations; the trained down-projection (6,144 $\times$ 2,240) returns the product to the residual stream. The profile reads the 6,144 numbers after the product and before the down-projection, the quantity \citet{geva2021} read as a key-value memory. The three matrices are the ones whose steps over training \citet{wilam2026} measures.

\begin{figure}[tp]
\centering
\renewcommand{\thefigure}{A.2}
\includegraphics[width=1\textwidth]{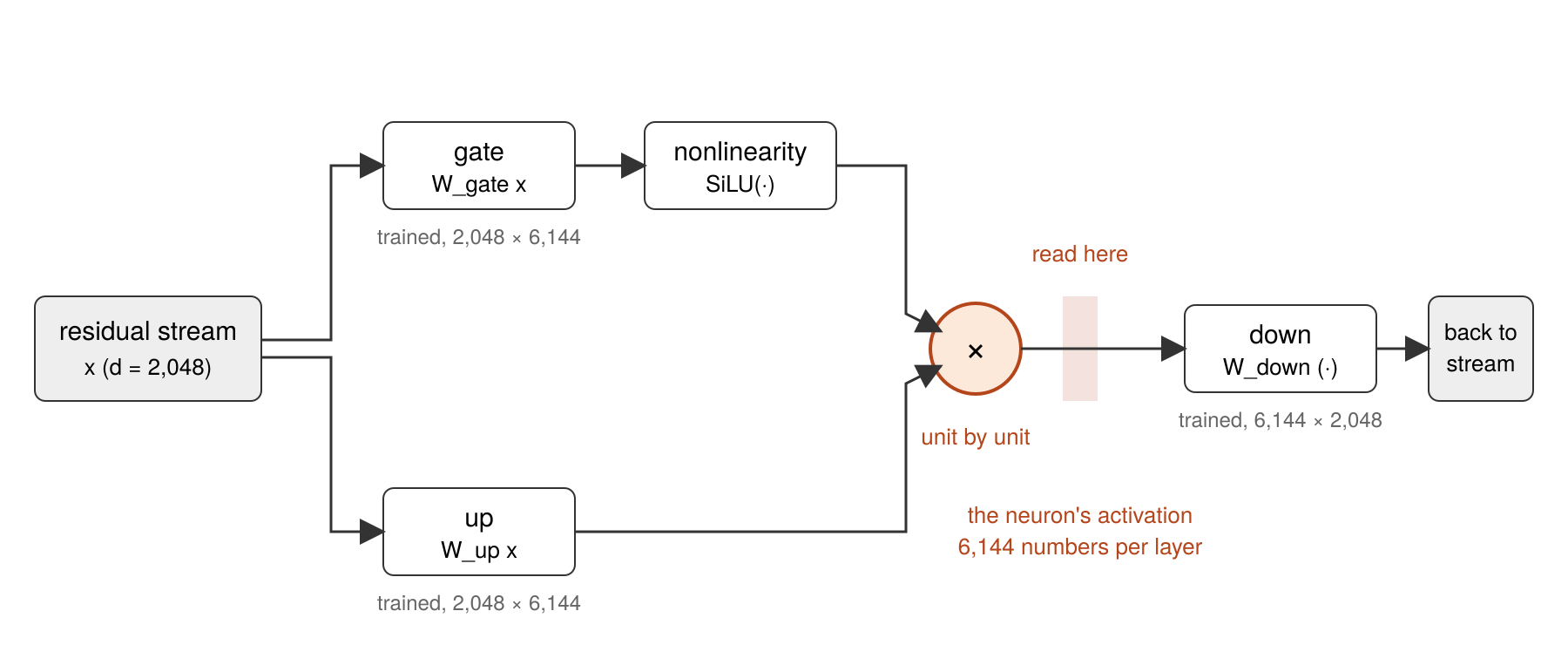}
\caption{The read point in one MLP block (SwiGLU): the neuron's activation is the gate branch after the nonlinearity times the up branch, taken before the down-projection.}
\label{fig:A.2}
\end{figure}

\subsection{The record}

\emph{What is stored.} Per reference, for every construct, both twins, every set and all 24 layers: per-core accumulators of the count and the mean raw activation per neuron, with the cell axis kept; the yardstick; and the loss at every offset of every prompt, which the ablation of \S4.4 reads. The accumulators take 27.6 GB a reference; the four per-prompt references are 107 GB each; 2,022 GB in all. What is stored per prompt is in \S3.2.

\emph{Per prompt.} The 31 constructs with per-prompt MLP activations were drawn so that the checks that need them license the population that is read: 15 constructs the design requires (the pilot's and the two-use pairs' members that pass the read-site gate) and 17 builtins, in the read population's proportion of one-piece and multi-piece tokens. One planned construct (\texttt{tuple}) is missing at all four references. Storing the neurons per prompt for every construct would have cost about six times as much: 890 GB across the four references instead of 147 (decimal gigabytes, the per-prompt neuron arrays alone; \texttt{RECORD\_SIZE.json}).

\subsection{The counts}

\begin{table}[H]
\centering
\scriptsize
\renewcommand{\arraystretch}{1.25}
\renewcommand{\thetable}{A.1}
\caption{The counts of the battery and the record.}
\label{tab:A.1}
\vspace{3pt}
\begin{tabular}{@{}>{\raggedright\arraybackslash}p{0.6667\dimexpr\textwidth-2\tabcolsep\relax}>{\raggedright\arraybackslash}p{0.3333\dimexpr\textwidth-2\tabcolsep\relax}@{}}
\toprule
object & count \\
\midrule
inventory: grammar entries + builtin names & 63 + 143 = 206 constructs \\\addlinespace[2.5pt]
built / not built & 188 / 18 \\\addlinespace[2.5pt]
pass the read-site gate (P$_1$) & 169 of 188 \\\addlinespace[2.5pt]
gate failures: structural / keyword tokens & 14 / 5 \\\addlinespace[2.5pt]
reference corpus & 253,398 parsed Python files \\\addlinespace[2.5pt]
prompts per A and per B set, per twin & 720 (8 cores $\times$ 90), all 188 constructs \\\addlinespace[2.5pt]
cores per construct & 8, from a family of 42--1,280 (median 210) \\\addlinespace[2.5pt]
prompts, tokens & 569,270; 33,271,350 \\\addlinespace[2.5pt]
tokens per prompt & mean 58.45; median 55; quartiles 28--81 \\\addlinespace[2.5pt]
yardstick cells & 20 (4 token classes $\times$ 5 fifths of the prompt) \\\addlinespace[2.5pt]
references recorded & 61: 35 pretraining, 24 annealing, 2 untrained \\\addlinespace[2.5pt]
storage & 27.6 GB of accumulators a reference; 107 GB at each of the four per-prompt references; 2,022 GB in all \\
\bottomrule
\end{tabular}
\end{table}

\subsection{Choices this realisation made, and their alternatives}

Each is a choice \S6 lists as a property of the instance, not of the method.
\begin{enumerate}[topsep=2pt,itemsep=1pt,parsep=0pt,partopsep=0pt,leftmargin=*]
\item \emph{Inventory:} the grammar entries and builtin names of Python 3.12. Alternatives: another language and its grammar; a curated subset.
\item \emph{Reference corpus:} public Python, since the model's training data was not released. Alternative: the training corpus, where released.
\item \emph{Cores and twins:} eight cores, shared by the two twins. Alternatives: more cores with fewer prompts each, which trades surface noise for core noise; twins with independent cores, which would make the twin unit carry both (\S4.8, artefact 1).
\item \emph{Read site:} the first token of the construct's node, on its MLP neurons. Alternatives: a later token of the construct; the residual stream; a read site that does not depend on the token piece. Each changes what is read, not the arithmetic.
\item \emph{Cells:} token class $\times$ fifth of the prompt, 20 cells. Other splits are possible, under the two requirements of \S3.3.
\end{enumerate}

The generator's build --- the grammars, the calibration tables, the companion mechanism, the matched control sets, the loss positions and the amendments between design and battery --- is documented with the code (Appendix E).

\section{The licence, construct by construct}

The checks of Table 4 at the final pretraining checkpoint (\texttt{pretrain\_iter\_0484865}, 2,034 B tokens); checks 2, 3, 7a and 7b after the re-run of 19 October (Appendix D). B.1 gives each check's spread over the constructs, B.2 the constructs the checks single out, B.3 the companion check and B.4 length sensitivity. The values for every construct are released as \texttt{TABLE\_B1.csv}, \texttt{TABLE\_B2.csv} and \texttt{TABLE\_B3.csv}.

\subsection{Each check over the constructs}

\begingroup
\scriptsize
\renewcommand{\arraystretch}{1.25}
\setlength{\LTcapwidth}{\textwidth}
\renewcommand{\thetable}{B.1}
\begin{longtable}{@{}>{\raggedright\arraybackslash}p{0.1400\dimexpr\textwidth-10\tabcolsep\relax}>{\raggedright\arraybackslash}p{0.3300\dimexpr\textwidth-10\tabcolsep\relax}>{\raggedright\arraybackslash}p{0.0600\dimexpr\textwidth-10\tabcolsep\relax}>{\raggedright\arraybackslash}p{0.1300\dimexpr\textwidth-10\tabcolsep\relax}>{\raggedright\arraybackslash}p{0.1800\dimexpr\textwidth-10\tabcolsep\relax}>{\raggedright\arraybackslash}p{0.1600\dimexpr\textwidth-10\tabcolsep\relax}@{}}
\caption{Each check over the constructs it ran on: the median, and the lowest and highest values with the constructs that hold them. The gate and checks 1, 4, 5 and 8 ran on all 188; their medians over the 169 that pass the gate are in brackets. Checks 2, 3, 7a and 7b ran on the 31 with per-prompt activations. Check 6 is in B.3.}\label{tab:B.1}\\
\toprule
check & what is read & ran on & median & lowest & highest \\
\midrule
\endfirsthead
\toprule
check & what is read & ran on & median & lowest & highest \\
\midrule
\endhead
\midrule
\multicolumn{6}{r}{\itshape continued} \\
\endfoot
\bottomrule
\endlastfoot
gate & share of prompts whose A and B read sites land on the same token piece (pass: 0.80 or above) & 188 & 169 pass, 19 fail (B.2) & 0.250 (\texttt{BinOp}, \texttt{Call}, \texttt{Compare}, \texttt{List}); among those that pass, 0.873 (\texttt{NamedExpr}) & 1.000 \\\addlinespace[2.5pt]
1 twin agreement & cosine between the two twins' profiles & 188 & 0.9861 (0.9889) & 0.7419 (\texttt{UnaryOp}); among the 169, 0.8771 (\texttt{None}) & 0.9980 (\texttt{NamedExpr}) \\\addlinespace[2.5pt]
2 convergence & prompts needed for a cosine of 0.99 with the profile on all 720, read on a grid of 8, 24, 56, 120 and 176 & 31 & 56 & 24 (\texttt{Delete}, \texttt{NotImplemented}, \texttt{With}, \texttt{ord}) & 120 (\texttt{For}, \texttt{If}, \texttt{ImportFrom}, \texttt{ResourceWarning}) \\\addlinespace[2.5pt]
3 length index & the length index of the profile & 31 & 0.75 & 0.41 (\texttt{IfExp}, \texttt{ListComp}) & 3.36 (\texttt{With}) \\\addlinespace[2.5pt]
3 flagged share & share of the construct's neurons flagged as length-sensitive & 31 & 0.024 & 0.002 (\texttt{IfExp}, \texttt{SetComp}) & 0.554 (\texttt{With}) \\\addlinespace[2.5pt]
4 re-weighting, cosine & cosine between the profile with the logged factors re-weighted and the base profile & 188 & 0.9095 (0.9135) & 0.6923 (\texttt{BinOp}); among the 169, 0.7852 (\texttt{Pass}) & 0.9943 (\texttt{Global}) \\\addlinespace[2.5pt]
4 re-weighting, twin units & the distance between the two over the construct's own twin distance (Appendix D) & 188 & 2.30 (2.35) & 0.71 (\texttt{Global}) & 4.94 (\texttt{bytearray}) \\\addlinespace[2.5pt]
5 calibration & factor by which the core-bootstrap intervals understate the twins' difference (1 is calibrated; above 1, too narrow) & 188 & 0.997 (1.005) & 0.45 (\texttt{callable}, \texttt{format}, \texttt{zip}) & 7.17 (\texttt{NotImplemented}) \\\addlinespace[2.5pt]
7a scrambled null & the real profile's norm at layer 12 over the scrambled-label null's 95th percentile (pass: above 1) & 31 & 10.66 & 7.22 (\texttt{With}) & 13.01 (\texttt{ord}) \\\addlinespace[2.5pt]
7b split-half null & the same over the split-half null's 95th percentile (pass: above 1) & 31 & 12.76 & 5.96 (\texttt{With}) & 15.83 (\texttt{NamedExpr}) \\\addlinespace[2.5pt]
8 yardstick floor & shift of the profile when the floor on $\sigma$$_0$ is set three ways, in twin units & 188 & 0.0039 (0.0041) & 0.0008 (\texttt{Attribute}, \texttt{UnaryOp}); among the 169, 0.0013 (\texttt{None}) & 0.0347 (\texttt{Nonlocal}) \\
\end{longtable}
\endgroup

All 31 pass checks 7a and 7b.

\subsection{The constructs the checks single out}

\emph{The gate.} 19 constructs fail, with the share of prompts on one piece in brackets: \texttt{AnnAssign} (0.617), \texttt{Assign} (0.406), \texttt{AsyncFunctionDef} (0.518), \texttt{Attribute} (0.375), \texttt{AugAssign} (0.375), \texttt{BinOp} (0.250), \texttt{BoolOp} (0.625), \texttt{Call} (0.250), \texttt{ClassDef} (0.529), \texttt{Compare} (0.250), \texttt{Dict} (0.371), \texttt{FunctionDef} (0.503), \texttt{List} (0.250), \texttt{Set} (0.375), \texttt{Slice} (0.500), \texttt{Starred} (0.500), \texttt{Subscript} (0.542), \texttt{Tuple} (0.371), \texttt{UnaryOp} (0.375). Of the 169 that pass, all but two are at 0.995 or above; the two are \texttt{NamedExpr} (0.873) and \texttt{Ellipsis} (0.992). Why they fail is in Appendix A.4.

\emph{Check 3.} Eight statements have more than 5\% of their neurons flagged as length-sensitive: \texttt{With} 55\%, \texttt{For} 49\%, \texttt{If} 47\%, \texttt{While} 36\%, \texttt{ImportFrom} 31\%, \texttt{Assert} 21\%, \texttt{Delete} 10\%, \texttt{Return} 8\%. B.4 reads them with the length held.

\emph{Check 4.} Four constructs sit within one twin unit of their base profile: \texttt{Global} (0.71), \texttt{UnaryOp} (0.74), \texttt{Continue} (0.77), \texttt{Attribute} (0.99). Two of them, \texttt{UnaryOp} and \texttt{Attribute}, fail the gate.

\emph{Check 5.} 33 constructs have intervals understated by more than 1.5, 30 of them among the 169 read: \texttt{NotImplemented} 7.17, \texttt{Exception} 5.44, \texttt{KeyboardInterrupt} 4.25, \texttt{EnvironmentError} 3.83, \texttt{ConnectionAbortedError} 3.76, \texttt{PendingDeprecationWarning} 3.68, \texttt{ModuleNotFoundError} 3.63, \texttt{ResourceWarning} 3.53, \texttt{IfExp} 3.24, \texttt{GeneratorExp} 2.86, \texttt{Pass} 2.86, \texttt{ListComp} 2.86, \texttt{SetComp} 2.72, \texttt{DictComp} 2.47, \texttt{Try} 2.20, \texttt{globals} 2.19, \texttt{AsyncFor} 2.12, \texttt{isinstance} 2.10, \texttt{iter} 2.09, \texttt{AsyncWith} 2.05, \texttt{Assert} 2.03, \texttt{Delete} 1.95, \texttt{Match} 1.95, \texttt{ord} 1.92, \texttt{hex} 1.85, \texttt{divmod} 1.85, \texttt{ascii} 1.77, \texttt{Return} 1.73, \texttt{memoryview} 1.68, \texttt{While} 1.67. The three that fail the gate: \texttt{ClassDef} 2.00, \texttt{AsyncFunctionDef} 1.97, \texttt{FunctionDef} 1.75.

\subsection{The companion check}

Check 6, on the eight constructs that can have a companion. The asymmetry is the difference between A's and B's share of the prompts that carry a companion; it is a count over prompt text, measured on the battery the model was run on, and has no layer.

\emph{Why two readings.} The record does not keep the boundary between a prompt's padding and its core. \emph{Whole} counts every line; \emph{pad-0} counts only the prompts drawn without padding, where the boundary is the first line and the count is exactly the generator's --- a third of each A set by design, 14--35\% of each B set. Companions that are the construct's own components, as the inventory lists them, are excluded; the re-weighting of \S4.1 excludes the same ones. The generator's audit uses a shorter list, which differs for four of the eight (\texttt{For}, \texttt{While}, \texttt{Try}, \texttt{Match}) and changes only \texttt{Match} materially: its largest difference is 0.186 here and 0.980 under the audit's list (both in \texttt{CHECK6\_COMPANIONS.json}).

\emph{Where the difference comes from, and where it can be removed.} Some of the difference is what the construct's grammar requires (every \texttt{match} case ends in \texttt{pass}; an \texttt{if} test is usually a comparison); some comes from how B was built (B's strings are f-strings; its sibling-role prompts are drawn from other constructs, the async variants among them). B's prompts are ordinary code with the token out of role and carry companions at a fifth of prompts or so, not near zero. Re-weighting B to A's companion shares needs per-prompt activations: for \texttt{For} and \texttt{With} the shares are attainable and the shift is measured (\S4.1); for \texttt{If} and \texttt{While} no weighting of the drawn prompts attains them (\texttt{While}'s B contains a comparison in every prompt, against a share of 0.69 in A); for \texttt{ClassDef}, \texttt{FunctionDef}, \texttt{Match} and \texttt{Try} no per-prompt activations exist.

\emph{Which neurons are named, and why the re-weighting is not adopted.} \texttt{With}'s recorded and re-weighted carrier sets overlap by 0.77 at the final checkpoint, below the band of the two conditions' twin ceilings (0.92--0.94), and likewise at the other trained references; for \texttt{For} the two sets agree within that band at every reference. The re-weighting rests on a fifth to a third of the prompts and lowers the twin ceiling at every reference (\texttt{For} by 0.05--0.12, \texttt{With} by 0.02--0.13), so the companion difference is stated as a restriction rather than corrected.

\begin{table}[t]
\centering
\scriptsize
\renewcommand{\arraystretch}{1.25}
\renewcommand{\thetable}{B.3}
\caption{Check 6, for the eight constructs: the concept norm at layer 12, and the asymmetry --- as an L2 norm over all companions and as the largest single one (max) --- counted over every line of every prompt (whole) and over the prompts drawn without padding (pad-0). The construct's own components, as the inventory lists them, are excluded.}
\label{tab:B.3}
\vspace{3pt}
\begin{tabular}{@{}>{\raggedright\arraybackslash}p{0.2500\dimexpr\textwidth-12\tabcolsep\relax}>{\raggedright\arraybackslash}p{0.1250\dimexpr\textwidth-12\tabcolsep\relax}>{\raggedright\arraybackslash}p{0.1250\dimexpr\textwidth-12\tabcolsep\relax}>{\raggedright\arraybackslash}p{0.1250\dimexpr\textwidth-12\tabcolsep\relax}>{\raggedright\arraybackslash}p{0.1250\dimexpr\textwidth-12\tabcolsep\relax}>{\raggedright\arraybackslash}p{0.1250\dimexpr\textwidth-12\tabcolsep\relax}>{\raggedright\arraybackslash}p{0.1250\dimexpr\textwidth-12\tabcolsep\relax}@{}}
\toprule
construct & concept norm (layer 12) & L2 whole & L2 pad-0 & max whole & max pad-0 & companions \\
\midrule
ClassDef & 53.06 & 1.064 & 1.311 & 0.388 & 0.646 & 33 \\\addlinespace[2.5pt]
For & 50.54 & 0.644 & 0.780 & 0.228 & 0.358 & 36 \\\addlinespace[2.5pt]
FunctionDef & 36.12 & 0.812 & 0.863 & 0.530 & 0.526 & 34 \\\addlinespace[2.5pt]
If & 44.93 & 0.919 & 0.989 & 0.678 & 0.688 & 35 \\\addlinespace[2.5pt]
Match & 47.65 & 0.395 & 0.464 & 0.186 & 0.261 & 32 \\\addlinespace[2.5pt]
Try & 70.75 & 0.572 & 0.729 & 0.266 & 0.524 & 33 \\\addlinespace[2.5pt]
While & 72.57 & 1.043 & 1.187 & 0.498 & 0.542 & 34 \\\addlinespace[2.5pt]
With & 55.46 & 0.693 & 1.030 & 0.353 & 0.667 & 36 \\
\bottomrule
\end{tabular}
\end{table}

\subsection{Length sensitivity under the yardstick}

For the eight length-sensitive statements of B.2, padding is added before the construct, so it moves the read site to a later fifth of the prompt in 54--98\% of prompts across the eight, and each padding level is then standardised in a different cell. Held in its unpadded cell, the length index falls by a median 26\% and the flagged share by 41\%; \texttt{Delete} and \texttt{Return} fall below 5\%, and six remain --- \texttt{With} 40\%, \texttt{For} 31\%, \texttt{If} 31\%, \texttt{While} 19\%, \texttt{ImportFrom} 16\%, \texttt{Assert} 14\%. The direction of the profile is largely kept: the lowest cosine between padding levels rises to 0.88--0.99 (\texttt{With} 0.79 $\rightarrow$ 0.97).

\section{Supporting readings for \S4}

\emph{Each item names the section it supports: C.1 the ablation and the yardstick (\S4.3, \S4.4); C.2 the running example over training (\S4.2, \S4.8); C.3 the map behind Figure 2 (\S4.5); C.4 the shared neurons (\S4.6); C.5 the probe (\S4.7).}

\subsection{The ablation comparison across yardsticks}

Each yardstick names a carrier set per construct; the set is ablated and the loss rise is read over the mean of 20 size-matched random sets, in nats. Medians over the 16 position-marked constructs, with the number of constructs above the random mean (sign test). \texttt{Try} is included: its carriers cost 5.95 nats more than a random set on a 0.088-nat baseline.

\begin{table}[t]
\centering
\scriptsize
\renewcommand{\arraystretch}{1.25}
\renewcommand{\thetable}{C.1}
\caption{Loss rise over the random mean, nats (constructs above, of 16).}
\label{tab:C.1}
\vspace{3pt}
\begin{tabular}{@{}>{\raggedright\arraybackslash}p{0.3333\dimexpr\textwidth-4\tabcolsep\relax}>{\raggedright\arraybackslash}p{0.3333\dimexpr\textwidth-4\tabcolsep\relax}>{\raggedright\arraybackslash}p{0.3333\dimexpr\textwidth-4\tabcolsep\relax}@{}}
\toprule
reference & 20-cell split & one pooled cell \\
\midrule
4.19 B tokens & +0.08 (13) & +0.25 (15) \\\addlinespace[2.5pt]
83.9 B & +0.27 (14) & +0.51 (14) \\\addlinespace[2.5pt]
2,034 B & +0.40 (14) & +0.73 (14) \\
\bottomrule
\end{tabular}
\end{table}

The 20-cell split gives the lower effect at every checkpoint, as expected: it excludes neurons that are large only because of where they stand, and ablation rewards size. The 20-cell figures are from a 20-mask run; \S4.4's are from the 200-mask run of the same experiment (\texttt{TABLE\_C3\_yardsticks.md}).

\emph{Why the comparison is with the random mean, in nats.} The random sets' spread is long-tailed at the two later references --- a few random sets hit load-bearing neurons --- and it keeps growing with more draws, so it does not settle into a usable unit (\S4.4).

\emph{Per construct.} Taken one at a time against 320 random sets, 2 of 16 constructs exceed every random set at 4.19 B tokens, 8 at 84 B and 8 at the end, and 2, 8 and 8 remain clear at 0.05 after a Holm correction. The floor of 1/321 is the resolution a 320-set control allows; the paper's claim is the one over constructs.

\subsection{\texttt{for} at four references}

Untrained, 4.19 B, 83.9 B and 2,034 B tokens. The running example's numbers over training, behind \S4.2; the raw norm falls from 4.19 B on while the standardised norm rises to 83.9 B and ends above its 4.19 B value, which is artefact 2 of \S4.8 seen on one construct. Raw norm 15.7 $\rightarrow$ 184.5 $\rightarrow$ 143.0 $\rightarrow$ 96.0. Standardised norm 30.2 $\rightarrow$ 239.6 $\rightarrow$ 270.9 $\rightarrow$ 251.5. Sparsity 0.390 $\rightarrow$ 0.471 $\rightarrow$ 0.538 $\rightarrow$ 0.505. Twin agreement 0.427 $\rightarrow$ 0.991 $\rightarrow$ 0.984 $\rightarrow$ 0.984. Depth 10.94 $\rightarrow$ 11.89 $\rightarrow$ 11.09 $\rightarrow$ 10.90. Twin ceiling 0.293 $\rightarrow$ 0.902 $\rightarrow$ 0.889 $\rightarrow$ 0.872. The two twins agree almost perfectly from the first checkpoint on.

\subsection{The full neuron map (Figure C.1)}

In Figure C.1, all 24 $\times$ 6,144 MLP neurons of OpenCoder-1.5B drawn in place at the end of pretraining, one row per layer, the neurons within a layer ordered by degree as in Figure 2, the order fixed once and reused.

\begin{figure}[tp]
\centering
\renewcommand{\thefigure}{C.1}
\includegraphics[width=1\textwidth]{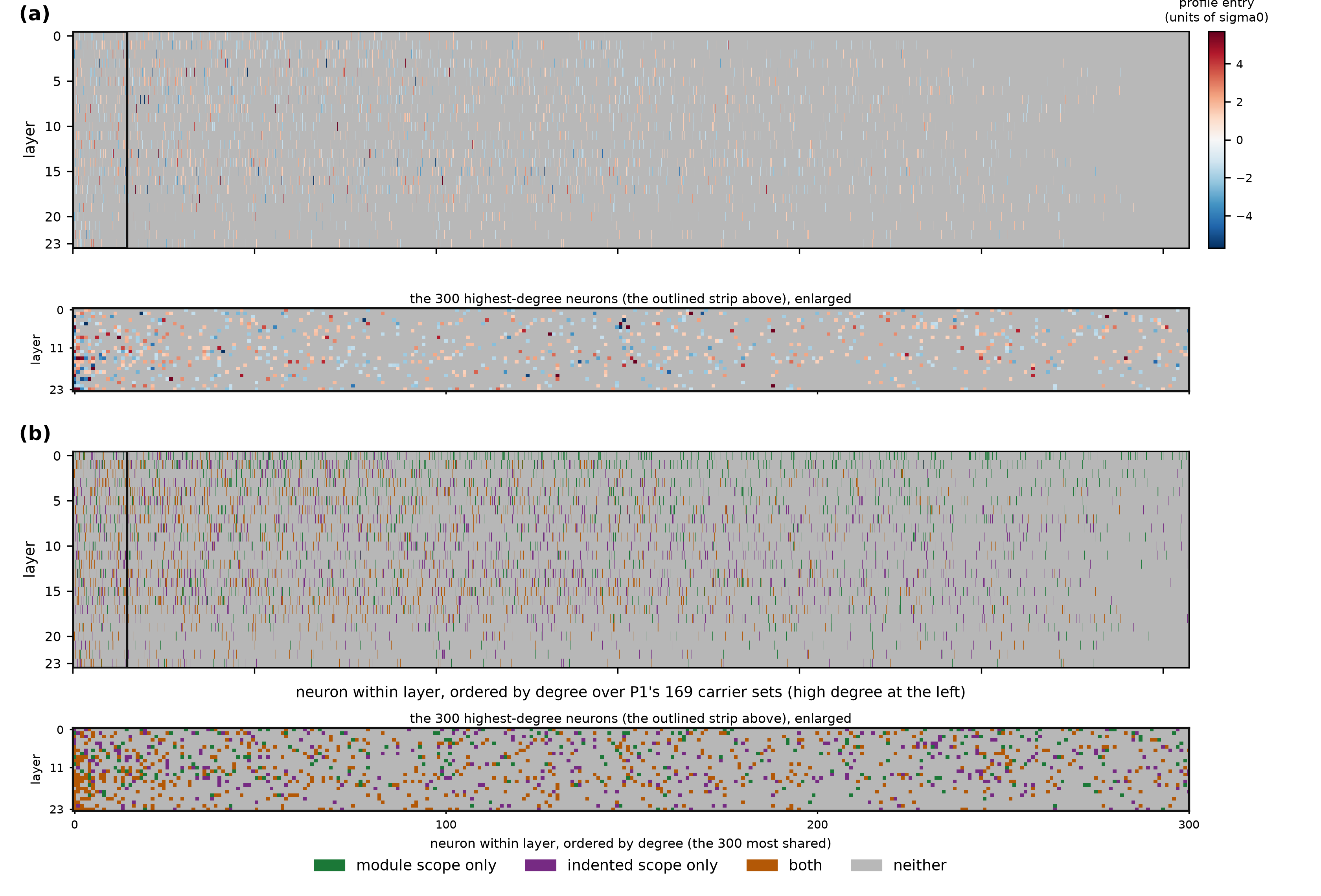}
\caption{Panel (a): the \texttt{for} loop's carrier set, each carrier in the colour of its entry $\pi$(A) $-$ $\pi$(B) in units of $\sigma$$_0$, every other neuron grey; an inset enlarges the 300 highest-degree neurons so that single entries can be read. Panel (b): one role, two tokens --- the indented \texttt{\textvisiblespace{}for} and the module-level \texttt{for}, two unrelated vocabulary entries; each neuron coloured by which of their two carrier sets it belongs to (module only, indented only, both, neither), with the same inset. What the map shows: the carriers are densest at the high-degree end and present at nearly every degree, falling to none among the lowest-degree neurons, and present in every layer; in panel (b) the shared neurons (both) and the scope-specific ones are interleaved everywhere, not grouped by layer or by degree (their overlap is given in \S4.5). What it does not show: individual entries outside the inset, or any layer pattern finer than presence in all 24.}
\label{fig:C.1}
\end{figure}

\subsection{Shared neurons by category (\S4.6)}

\emph{What was clustered, and why.} Two constructs share neurons when their carrier sets overlap. The question is whether the constructs that share most fall into the grammar's categories or into the inventory's. The overlaps of P$_1$'s 14,196 construct pairs at the last checkpoint were clustered into 2 to 32 clusters, and each category was scored for purity --- the share of a cluster's members that belong to the category --- against 1,000 shuffles of the labels.

\begin{table}[t]
\centering
\scriptsize
\renewcommand{\arraystretch}{1.25}
\renewcommand{\thetable}{C.2}
\caption{Categories recovered by the clustering (p < 0.001 unless noted).}
\label{tab:C.2}
\vspace{3pt}
\begin{tabular}{@{}>{\raggedright\arraybackslash}p{0.3333\dimexpr\textwidth-4\tabcolsep\relax}>{\raggedright\arraybackslash}p{0.3333\dimexpr\textwidth-4\tabcolsep\relax}>{\raggedright\arraybackslash}p{0.3333\dimexpr\textwidth-4\tabcolsep\relax}@{}}
\toprule
category & constructs & purity \\
\midrule
grammar element against builtin name & --- & recovered \\\addlinespace[2.5pt]
exception classes & 69 & 1.000 \\\addlinespace[2.5pt]
warning classes & 12 & 1.000 \\\addlinespace[2.5pt]
builtin functions & 43 & 0.86 \\\addlinespace[2.5pt]
builtin types & 26 & 0.73 \\\addlinespace[2.5pt]
comprehensions & 4 & 1.000 (p = 0.004) \\
\bottomrule
\end{tabular}
\end{table}

\emph{Not recovered:} statement against expression, compound statements, binding statements, composites and their obliged parts.

\emph{Other readings.} The highest-sharing pairs across categories are \texttt{Delete}--\texttt{delattr} (0.200) and \texttt{Global}--\texttt{globals} (0.094). Keyword tokens sit at a median depth of 10.98 layers and builtins at 11.97, with indistinguishable concentration (sparsity 0.666 against 0.673). Under one pooled cell, 38 of 189 purity tests (7 cuts $\times$ 6 categories, Holm-corrected) are significant, all of them inventory categories (\texttt{POOLED\_SETS\_45.json}).

\subsection{The probe in full (\S4.7)}

The area under the curve: a median 1.000 at every trained reference, 145--148 of 164 constructs at or above 0.99. Two roles of one token (\texttt{for} as loop against \texttt{for} in a comprehension) read 0.68--0.75 untrained and 0.999 at the median from the first checkpoint on. The probe's margin, the signed distance of a held-out prompt from the boundary in the probe's own units, moves: a median 3.2 untrained, 13.8 at 4.19 B, 16.5 at 83.9 B and 14.8 at the end, against a control margin near zero throughout. It rises to 83.9 B for 128 of 164 constructs and falls again to the end for 149 of 164 (paired sign tests, p < 10$^-$$^1$$^2$), ending 7\% above its value at 4.19 B --- a turn at the same point where the raw size of the response begins to fall \citep[\S4.4]{wilam2026}.

\section{Reproducibility}

Every figure in \S3 and \S4 is read from the record or from the battery's own files as stored, not estimated; where a figure is an approximation the text says so. Battery hash \texttt{570b42db\ldots{}bcbc1ad}. The release: the dataset \texttt{piotrwilam/AtlasOpenCoder}, tag \texttt{v1}, DOI 10.57967/hf/11472 --- the battery in full, the carrier sets at every reference, the licence tables, the readings behind the figures, and a worked example of the record and of the profiles for \texttt{for} --- and the repository \texttt{piotrwilam/AtlasCalibratedConceptProfiles} (the generator, the recorder, the checks and every script of Table D.1). The dataset's \texttt{MANIFEST\_PAPER1.md} lists every number in this paper with the file and field it is read from and the script that rebuilds it.

\emph{Where each number comes from.} Each table and figure, the stored file it is read from, and the script that wrote the version in this paper. Files are in \texttt{figures/} and \texttt{figures/data/}; scripts in \texttt{experiments/}.

\begingroup
\scriptsize
\renewcommand{\arraystretch}{1.25}
\setlength{\LTcapwidth}{\textwidth}
\renewcommand{\thetable}{D.1}
\begin{longtable}{@{}>{\raggedright\arraybackslash}p{0.3333\dimexpr\textwidth-4\tabcolsep\relax}>{\raggedright\arraybackslash}p{0.3333\dimexpr\textwidth-4\tabcolsep\relax}>{\raggedright\arraybackslash}p{0.3333\dimexpr\textwidth-4\tabcolsep\relax}@{}}
\caption{Where each figure and table comes from.}\label{tab:D.1}\\
\toprule
in the paper & file & script \\
\midrule
\endfirsthead
\toprule
in the paper & file & script \\
\midrule
\endhead
\midrule
\multicolumn{3}{r}{\itshape continued} \\
\endfoot
\bottomrule
\endlastfoot
Tables 1--3 (the factors, the nine checks, the six choices) & --- (text) & --- \\\addlinespace[2.5pt]
Table 4 (the licence as run) & \texttt{LICENCE\_TABLE.json}, \texttt{APPENDIX\_B1\_checks.md} & \texttt{licence\_table.py}, \texttt{licence\_tables.py} \\\addlinespace[2.5pt]
Appendix B.1--B.2, the gate and checks 1, 4, 5, 8 & \texttt{TABLE\_B1.csv} & \texttt{licence\_tables.py} \\\addlinespace[2.5pt]
Appendix B.1 and D, check 4 in twin units & \texttt{CHECK4\_RELATIVE.\{csv,json\}} & \texttt{check4\_relative.py} \\\addlinespace[2.5pt]
Appendix B.1--B.2, checks 2, 3, 7a, 7b & \texttt{TABLE\_B2.csv}; \texttt{RECHECK\_PER\_PROMPT.json} (check 2's prompt counts) & \texttt{licence\_tables.py} \\\addlinespace[2.5pt]
Appendix B.1, the two null ratios & \texttt{NULL\_RATIOS.json} (the layer-12 real norm, each null's p95 and the ratio) & \texttt{licence\_tables.py} \\\addlinespace[2.5pt]
Appendix B.1--B.2, the gate & \texttt{GATE\_SHARES.json} (the four shares per construct and their minimum) & \texttt{licence\_tables.py} \\\addlinespace[2.5pt]
\S2.6, \S3.3 and A.8, the twin ceilings & \texttt{TWIN\_CEILINGS.json} (every construct at all 61 references, both spaces) & \texttt{twin\_ceilings.py} \\\addlinespace[2.5pt]
Appendix B.3 (companion) & \texttt{CHECK6\_COMPANIONS.json} (\texttt{table\_b3.rows}), \texttt{TABLE\_B3.csv} & \texttt{check6\_companions.py}, \texttt{licence\_tables.py} \\\addlinespace[2.5pt]
Appendix B.4 (length, band held) & \texttt{LENGTH\_BAND\_HELD.json} & \texttt{band\_held.py} \\\addlinespace[2.5pt]
Figure 1 (the method in six steps) & \texttt{FIG1\_method.\{png,svg,json\}} & \texttt{figure\_method.py} --- a schematic; nothing in it is measured \\\addlinespace[2.5pt]
Figure 2 (\texttt{for}'s carriers in place) & \texttt{FIG1\_carriers\_in\_place.\{png,svg,json\}} & \texttt{figure1\_small.py} \\\addlinespace[2.5pt]
Figure A.1 (three prompts from the battery) & \texttt{phase\_battery/twin1/For\_A.jsonl.gz} & none: quoted from the battery files \\\addlinespace[2.5pt]
Figure A.2 (the neuron read) & \texttt{FIG\_neuron\_swiglu.\{png,svg\}} & none: a schematic, drawn by hand; it shows no data \\\addlinespace[2.5pt]
Figure C.1 (the full neuron map) & \texttt{FIGC1\_neuron\_map.\{png,svg,json\}} & \texttt{figure\_c1.py} \\\addlinespace[2.5pt]
\S3.1 core families & \texttt{CORE\_COUNTS.\{json,md\}} & \texttt{core\_counts.py} \\\addlinespace[2.5pt]
\S3.1 B's sub-sets & \texttt{B\_SUBTYPES.\{json,md\}} & \texttt{b\_subtypes.py} \\\addlinespace[2.5pt]
\S4.2 resolution & \texttt{B10\_SPACING\_ADJUSTED.json} & \texttt{b10\_fit.py} \\\addlinespace[2.5pt]
\S4.3 coarser yardsticks & \texttt{TURNOVER\_CLASSES\_ONLY.json} & \texttt{b10.py} \\\addlinespace[2.5pt]
\S4.4 ablation & \texttt{ABLATION\_NATS.json}, \texttt{TABLE\_D1\_ablation.\{md,csv\}} & \texttt{ablation\_ranks.py} for the version in the paper (it writes the rank and exact-p columns and removes control units), reading \texttt{ablation\_full.py}'s output; \texttt{ablation\_stable.py} and \texttt{ablation\_nats.py} also write \texttt{TABLE\_D1\_ablation} and give different numbers \\\addlinespace[2.5pt]
\S4.5 two roles of one token & \texttt{TWO\_USE\_SHARING.json} & \texttt{atlas.py b5} \\\addlinespace[2.5pt]
\S4.5 one role, two tokens & \texttt{SCOPE\_OVERLAP.json} & \texttt{atlas.py b4} \\\addlinespace[2.5pt]
\S4.6 and C.4 shared neurons by category & \texttt{FAMILIES.json} & \texttt{atlas.py b6} \\\addlinespace[2.5pt]
\S4.6 the pooled-cell replication & \texttt{POOLED\_SETS\_45.json} & \texttt{pooled\_sets\_169.py} \\\addlinespace[2.5pt]
\S4.7 probe margin & \texttt{PROBE\_MARGINS.json}, \texttt{TABLE\_probe\_margins.\{md,csv\}} & \texttt{probe\_margins.py}, reading \texttt{PROBE.json} from \texttt{b9\_b10.py} \\\addlinespace[2.5pt]
\S4.8 the three artefacts & \texttt{LICENCE\_TABLE.json} (1, the twin unit against the core bootstrap), \texttt{FOR\_EXAMPLE.json} (2, standardised against raw size), \texttt{FAMILIES.json} (3, the layer of largest response) & \texttt{licence\_table.py}, \texttt{atlas.py b3}, \texttt{atlas.py b6} \\\addlinespace[2.5pt]
Appendix C.1 (yardsticks, in nats) & \texttt{C3\_YARDSTICK\_NATS.json}, \texttt{TABLE\_C3\_yardsticks.\{md,csv\}} & \texttt{c3\_yardstick\_nats.py} \\\addlinespace[2.5pt]
Appendix C.2 (\texttt{for} at four references) & \texttt{FOR\_EXAMPLE.json} & \texttt{atlas.py b3} \\\addlinespace[2.5pt]
Appendix C.5 (the probe in full) & \texttt{PROBE.json}, \texttt{PROBE\_TWO\_ROLE.json} & \texttt{b9\_b10.py} \\\addlinespace[2.5pt]
Appendix D hashes & \texttt{PROVENANCE.md}, \texttt{IRREPLACEABLE.json} & \texttt{build\_provenance.py}, \texttt{irreplaceable.py} \\\addlinespace[2.5pt]
Appendix D, the licence re-run & \texttt{LICENCE\_TABLE.json} (as published), \texttt{LICENCE\_TABLE\_CORRECTED.json} (the 19 Oct re-run) & \texttt{licence\_table.py} \\\addlinespace[2.5pt]
\S3.5 watched numbers & \texttt{record\_GATE\_111.json} & \texttt{profilebattery/validation/watched.py} \\\addlinespace[2.5pt]
A.6 storage & \texttt{RECORD\_SIZE.json} & \texttt{record\_size.py} \\
\end{longtable}
\endgroup

Every numbered figure and table in the paper, and every section of \S4 and Appendix C that reports a measured number, has a row above; each named file was checked to exist in the release tree before the row was written. Two rows name no script, and say why: Figure A.1 quotes three prompts straight from the battery files, and Figure A.2 is a hand-drawn schematic that shows no data.

\emph{The untrained reference.} Its random weights do not regenerate across library versions, and the library versions used for its two recordings were not stored; the recorded profiles are kept, and a reproducible seed is owed.

\emph{Licence re-run.} Four checks were re-run on 19 Oct 2026 after a fault in how per-prompt vectors were joined to their prompts was found and fixed (the files are stored in the recorder's length-sorted order, not in jsonl row order). Convergence was unchanged at the median (56 prompts for a cosine of 0.99). Length sensitivity was lower than first reported (median index 0.7455 against 1.2142; flagged share 2.4\% against 3.9\%; the two earlier values are from the pre-fix run's report and are not in a released artefact, since \texttt{LICENCE\_TABLE.json} kept only the length levels and the minimum pair cosine). The re-run also drew 20 null replicates per construct, against 8 before. Both nulls fell. \emph{Their centre} --- the null profile's norm at layer 12, in the profile's own units, the median over replicates and then over the 31 constructs --- went from 4.86 to 4.57 for the scrambled null and from 5.27 to 4.64 for the split-half. \emph{Their tail} --- the 95th percentile, against which \S4.1's margin is taken --- fell further for the split-half: from 6.42 to 6.13 ($-$4.5\%) for the scrambled null and from 6.45 to 5.30 ($-$18\%) for the split-half. The margin above the tail, the real profile's norm over the null's 95th percentile (\S4.1), therefore rose: from 10.1$\times$ to 10.7$\times$ for the scrambled null and from 9.6$\times$ to 12.8$\times$ for the split-half. It had been understated, not overstated; quoting the centres alone would understate how much more the split-half margin moved.

\emph{Check 4 in twin units.} The licence table kept check 4's cosine and not its distance, so Table B.1's twin-unit row is rebuilt from both arms' per-core, per-cell accumulators: the median over the 188 is 2.30 against the licence suite's 2.34, and 4 of 188 sit within one twin unit, as \S4.1 states (\texttt{check4\_relative.py}, \texttt{CHECK4\_RELATIVE.csv}). A second route, from the stored cosines through the chord identity d = $\surd$(2(1 $-$ cos)), holds per pair but not between medians: on check 1, where both forms are stored, it is within 5\% for 90\% of constructs and at worst 18.6\% off, and it gives a median of 2.25 (quartiles 1.82--3.49), with 7 of 188 within one twin unit.

\section{Applying the method}

\emph{For a reader who wants to run the method on another model or language. The steps are those of \S2; the costs are this realisation's.}

\begin{enumerate}[topsep=2pt,itemsep=1pt,parsep=0pt,partopsep=0pt,leftmargin=*]
\item \emph{Inputs.} A model with at least one checkpoint (one checkpoint gives a reading; a series gives readings across training). An inventory of named constructs fixed before the model is read. A reference corpus in the language, for the declared rates and for set C. A parser or tagger that decides each construct's role, with a surface token per construct. Model access that returns activations at a named site. (\S6, \emph{What changes in another setting}.)
\item \emph{Declare the distribution.} List the factors around a construct (identifiers, literals, scope and nesting, companions, length, surrounding code) and measure each one's rate on the reference corpus. Here: codeparrot-clean, 253,398 parsed Python files.
\item \emph{Generate the battery.} For each construct, set A (the construct in its role) and set B (its token out of role), the same companions in both, drawn twice --- the twins --- from eight cores per construct; set C from the reference corpus. Log every draw; hash the battery. Here: 188 constructs, 720 prompts per set and twin, 569,270 prompts in all.
\item \emph{Gate the read site.} Keep a construct only if its A and B prompts put the read token on the same piece in at least 80\% of prompts (Appendix A.4). Here: 169 of 188 pass.
\item \emph{Record.} Run the model on the battery at each checkpoint; store per-core means of every neuron's activation, with the cell axis kept, so the yardstick and the cut can be changed later; store per-prompt activations for a subset if checks 2, 3, 7a and 7b are wanted. Here: about 28 GB of per-core means per reference, and 79 GB more at a reference with per-prompt arrays (36 GB for the neurons of 31 constructs, 43 GB for the residual stream of 164).
\item \emph{Compute the calibrated concept profile.} Per neuron: the mean on A minus the mean on B, each in the neuron's own spread on C in the same cell; the twin difference gives the noise unit.
\item \emph{License.} Run the nine checks of Table 2; the two nulls must pass; report the other seven (Table 4 is the form).
\item \emph{Read.} The profile, the carrier set and its twin ceiling at each checkpoint; across checkpoints, the resolution first (\S2.6), then any date or comparison in twin units.
\item \emph{Check by removal.} Ablate the carrier set and a random set of the same size, and read the loss at the construct's positions (\S4.4).
\end{enumerate}

\emph{Code.} Five stages, each a script in \texttt{experiments/} of the repository \texttt{piotrwilam/AtlasCalibratedConceptProfiles}.

\begin{table}[t]
\centering
\scriptsize
\renewcommand{\arraystretch}{1.25}
\renewcommand{\thetable}{E.1}
\caption{The five stages, their scripts and their outputs.}
\label{tab:E.1}
\vspace{3pt}
\begin{tabular}{@{}>{\raggedright\arraybackslash}p{0.5000\dimexpr\textwidth-4\tabcolsep\relax}>{\raggedright\arraybackslash}p{0.2500\dimexpr\textwidth-4\tabcolsep\relax}>{\raggedright\arraybackslash}p{0.2500\dimexpr\textwidth-4\tabcolsep\relax}@{}}
\toprule
stage & script & output \\
\midrule
1. inventory & \texttt{split\_p1.py} & \texttt{inventory/inventory\_v2.json}, \texttt{validation/battery/SPLIT.json} \\\addlinespace[2.5pt]
2. corpus calibration (the declared list $\nu$) & \texttt{build\_calibration.py} & \texttt{calibration/*.json} \\\addlinespace[2.5pt]
3. battery & \texttt{build\_battery.py} & \texttt{phase\_battery/}, \texttt{CONTENT\_HASHES.json} \\\addlinespace[2.5pt]
4. recording & \texttt{run\_s15.py}, \texttt{run\_s16.py} (\texttt{run\_colab.py} for a hosted GPU) & \texttt{record/<ref>/} \\\addlinespace[2.5pt]
5. licence & \texttt{licence\_table.py}, \texttt{licence\_tables.py} & \texttt{LICENCE\_TABLE.json}, Tables B.1--B.3 \\
\bottomrule
\end{tabular}
\end{table}

The stage table and its scripts are released with the repository; a minimal example, running stages 3--5 and the profile for one construct on one small model, will follow.

\end{document}